\documentclass[lettersize,journal]{IEEEtran}
\usepackage{amsmath,amsfonts}
\usepackage{amssymb}
\usepackage{algorithmic}
\usepackage{algorithm}
\usepackage{array}
\usepackage{subfigure}
\usepackage{textcomp}
\usepackage{stfloats}
\usepackage{url}
\usepackage{verbatim}
\usepackage{graphicx}
\usepackage{cite}
\usepackage{booktabs}  
\usepackage{multirow}
\usepackage{caption}
\usepackage{hyperref}
\usepackage{tabularx}
\usepackage{diagbox}
\usepackage{bm}
\usepackage{breakurl}
\usepackage{enumerate}
\usepackage{verbatim}
\usepackage{float}
\usepackage{siunitx}
\usepackage{mdframed}
\usepackage{adjustbox}
\usepackage{cleveref}
\usepackage{authblk}
\usepackage{color}
\usepackage{verbatim}
\usepackage{float}
\usepackage{siunitx}
\usepackage{mdframed}
\usepackage{cleveref}
\usepackage{xurl}
\usepackage{soul}
\usepackage{tabularx}
\usepackage{tikz}
\usetikzlibrary{shapes.geometric, arrows}
\usepackage{layouts}
\usepackage{array}
\usepackage{wasysym}  % symbols
\usepackage{graphicx}

\newcolumntype{s}{>{\hsize=.3\hsize}X}
\newcolumntype{x}{>{\hsize=.5\hsize}X}

\newcommand{\Eqref}[1]{Eq. \eqref{#1}}

\newcommand{\veca}{\boldsymbol{a}}

\newcommand{\vecb}{\mathbf{b}}

\newcommand{\vece}{\mathbf{e}}
\newcommand{\vecF}{\mathbf{F}}

\newcommand{\vecg}{\mathbf{g}}

\newcommand{\vecq}{\mathbf{q}}

\newcommand{\vecu}{\mathbf{u}}

\newcommand{\vecx}{\mathbf{x}}

\newcommand{\vecdelta}[1]{\bm{\delta}_{#1}}

\newcommand{\vecmu}{\bm{\mu}}
\newcommand{\vecomega}{\bm{\omega}}
\newcommand{\vecrho}[1]{\bm{\rho}_{#1}}
\newcommand{\hatvecrho}[1]{\hat{\bm{\rho}}_{#1}}
\newcommand{\vecxi}{\bm{\xi}}

\newcommand{\matA}{\mathbf{A}}
\newcommand{\matB}{\mathbf{B}}

\newcommand{\matI}{\mathbf{I}}
\newcommand{\matJ}{\mathbf{J}}
\newcommand{\matK}{\mathbf{K}}
\newcommand{\matM}{\mathbf{M}}
\newcommand{\matN}{\mathbf{N}}
\newcommand{\matP}{\mathbf{P}}
\newcommand{\matR}{\bm{\mathit{R}}}
\newcommand{\matU}{\mathbf{U}}

\newcommand{\matX}{\mathbf{X}}

\newcommand{\inertia}{\mathbf{J}}
\newcommand{\inertiaload}{\mathbf{J}_{L}}
\newcommand{\loadmass}{m_{L}}
\newcommand{\half}{\frac{1}{2}}

\newcommand{\angvel}{\mathbf{\Omega}}
\newcommand{\angacc}{\dot{\angvel}}

\newcommand{\robotpos}[1]{\vecx_{#1}}
\newcommand{\robotrot}[1]{\matR_{#1}}

\newcommand{\robotvel}[1]{\dot{\vecx}_{#1}}
\newcommand{\robotacc}[1]{\ddot{\vecx}_{#1}}
\newcommand{\robotangvel}[1]{\angvel_{#1}}
\newcommand{\robotangacc}[1]{\angacc_{#1}}

\newcommand{\tensiondes}[1]{\vecmu_{#1,des}}
\newcommand{\loadpos}{\vecx_{L}}
\newcommand{\loadposdes}{\vecx_{L,des}}
\newcommand{\loadrot}{\matR_{L}}
\newcommand{\loadrotdes}{\matR_{L,des}}

\newcommand{\loadvel}{\dot{\vecx}_{L}}
\newcommand{\loadveldes}{\dot{\vecx}_{L,des}}
\newcommand{\loadacc}{\ddot{\vecx}_{L}}
\newcommand{\loadaccdes}{\ddot{\vecx}_{L,des}}
\newcommand{\loadangvel}{\angvel_{L}}
\newcommand{\loadangveldes}{\angvel_{L,des}}
\newcommand{\loadangacc}{\angacc_{L}}
\newcommand{\loadangaccdes}{\angacc_{L,des}}

\newcommand{\cablevec}[1]{\vecxi_{#1}}

\newcommand{\cablevel}[1]{\vecomega_{#1}}
\newcommand{\cableveldes}[1]{\vecomega_{#1,des}}
\newcommand{\cabledotvec}[1]{\dot{\vecxi}_{#1}}

\newcommand{\cableddotvec}[1]{\ddot{\vecxi}_{#1}}

\newcommand{\inputforce}[1]{\vecu_{\mathit{#1}}}
\newcommand{\inputperp}[1]{\vecu_{\mathit{#1}}^{\perp}}
\newcommand{\inputpara}[1]{\vecu_{\mathit{#1}}^{\lVert}}

\newcommand{\realnum}[1]{\mathbb{R}^{#1}}
\newcommand{\SOthree}{SO(3)}

\newcommand{\sumn}[1]{\sum_{#1=1}^n}
\newcommand{\norm}[1]{\left\lVert#1\right\rVert}

\newcommand{\twonorm}[1]{\left\lVert#1\right\rVert_2}
\newcommand{\prths}[1]{\left(#1\right)}

\newcommand{\crbrc}[1]{\left\{#1\right\}}

\DeclareMathOperator*{\argmin}{arg\,min}

\newcommand{\worldf}{\mathcal{I}}
\newcommand{\robotf}[1]{\mathcal{B}_{#1}}
\newcommand{\loadf}{\mathcal{L}}
\newcommand{\axis}[2]{\mathbf{e}_{#1}^{#2}}

\begin{document}

\title{\LARGE \bf RotorTM: A Flexible Simulator for Aerial Transportation\\ and Manipulation }

\author{
    Guanrui Li, Xinyang Liu, and Giuseppe Loianno
\thanks{The authors are with the New York University, Tandon School of Engineering, Brooklyn, NY 11201, USA. {\tt\footnotesize email: \{lguanrui, liuxy, loiannog\}@nyu.edu}.}
\thanks{This work was supported by the NSF CPS Grant CNS-2121391, the NSF CAREER Award 2145277, the Technology Innovation Institute, Qualcomm Research, Nokia, and NYU Wireless. Giuseppe Loianno
serves as a consultant for the Technology Innovation Institute. This arrangement has been reviewed and approved by New York University in accordance with its policy on objectivity in research.}
\thanks{The authors acknowledge Manling Li, and Kelsey Fontenot for their help and support on this research and experiments.}
}

\markboth{IEEE Transactions on Robotics, VOL.~XX, NO.~X, December 2023}%
{Shell \MakeLowercase{\textit{et al.}}: Bare Demo of IEEEtran.cls for IEEE Journals}

\makeatletter
\g@addto@macro\@maketitle{
\setcounter{figure}{0}
\centering
    \includegraphics[width=\textwidth]{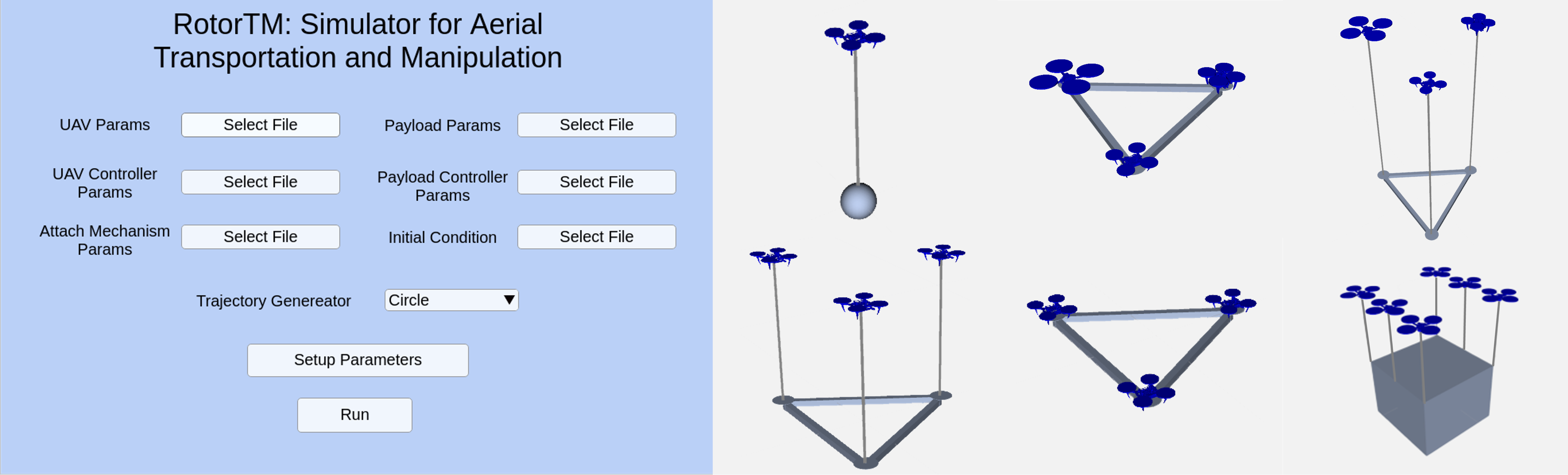}
	\captionof{figure}{Sample systems and user interface of the proposed RotorTM simulator.\label{fig:sample-system}}
}

\makeatother

\IEEEpubid{0000--0000/00\$00.00~\copyright~2023 IEEE}

\maketitle

\begin{abstract}
Low-cost autonomous Micro Aerial Vehicles (MAVs) have great potential to help humans by simplifying and speeding up complex tasks, such as construction, package delivery, and search and rescue. These systems, which may consist of single or multiple vehicles, can be equipped with passive connection mechanisms such as rigid links or cables for transportation and manipulation tasks. However, these systems are inherently complex. They are often underactuated and evolve in nonlinear manifold configuration spaces. In addition, the complexity escalates for systems with cable-suspended load due to the hybrid dynamics that vary with the cables' tension conditions. This paper presents the first aerial transportation and manipulation simulator incorporating different payloads and passive connection mechanisms with full system dynamics, planning, and control algorithms. Furthermore, it includes a novel general model accounting for the transient hybrid dynamics for aerial systems with cable-suspended load to closely mimic real-world systems. Comparisons between simulations and real-world experiments with different vehicles' configurations show the fidelity of the simulator results with respect to real-world settings. The experiments also show the simulator's benefit for the rapid prototyping and transitioning of aerial transportation and manipulation systems to real-world deployment.
\end{abstract}


\section*{Supplementary material}
\textbf{Video}: \url{https://youtu.be/jzfEVQ3qlPc}

\textbf{Code}: \url{https://github.com/arplaboratory/RotorTM}

\section{Introduction} \label{sec:intro}
\IEEEPARstart{L}{ow-cost} autonomous Micro Aerial Vehicles (MAVs) endowed with manipulation mechanisms have the potential to help humans in a wide range of complex and dangerous tasks such as construction~\cite{lindseyconstruction2012}, transportation and delivery~\cite{barmpounakis2016ijtst}, and inspection~\cite{angelsensors2019}. In construction scenarios, a MAV team can cooperatively transport construction materials from the ground to the upper floors, speeding up the construction process. Similarly, MAVs can speed up immediate humanitarian missions or emergency medical care deliveries in urban settings, which can often be delayed by ground traffic during rush hour, by exploiting the free ``highway" in the air. The tasks above require endowing aerial robots with the capability of transporting or manipulating objects. 

Researchers have employed both active and passive manipulation mechanisms for aerial transportation and manipulation~\cite{ollero2021TRO}. They used active mechanisms like actuated robot arms~\cite{jongseok2020icra,ouyang2021aerialmanipulator}, gripper~\cite{Mellinger2012CooperativeQuadrotors,pounds2011icra} and passive mechanisms like magnets~\cite{LoiannoRAL2018,rik2017ssrr}, spherical joints~\cite{tagliabue2019ijrr,Loiannodesert2018} and cables~\cite{guanrui2021ral,Klausen2020TCST}. Passive mechanisms require lower energy since they do not need extra actuation. In addition, they are generally lighter than active solutions. Therefore, passive mechanisms can potentially guarantee better flight endurance compared to active solutions. Moreover, they also have the advantage of being lower-cost and easier to design. Hence, passive attach mechanisms can benefit large-scale deployment of aerial payload transportation and manipulation in the real world. 
\IEEEpubidadjcol

However, by not directly actuating the load and presenting a configuration space that evolves on complex nonlinear manifolds, these systems require additional efforts to design control and planning strategies. It is usually dangerous, expensive, and time-consuming to directly design and test algorithms on robots or collect data in the real world. For these reasons, simulation tools provide cheap and fast access for researchers to collect data, prototype systems, and test algorithms. It is challenging to simulate aerial transportation and manipulation with MAVs that have passive mechanisms for the aforementioned difficulties. Furthermore, the hybrid dynamics introduced by the suspended cables create additional challenges that have to be properly modeled and accounted for. When the cables switch from slack to taut condition, the payload's and robots' velocity will transition to new values due to the collision along the cable direction.

Researchers have been spending efforts to develop their numerical simulation in MATLAB or Python~\cite{wang2018icra, wu2014geometric, Lee2018, RSS2013Koushil, cruz2015acc, pratik2020icra, rastgoftar2018calm, geng2022jais} to resolve the above challenges and validate control and planning algorithm design on aerial transportation and manipulation. However, these simulations are customized to address specific research problems, and most did not release the corresponding source code. Other researchers employ open-source generic simulators like~\cite{Furrer2016,okoli2019report}. However, adapting these environments to simulate MAVs transporting a payload via cables is challenging, especially considering the hybrid dynamics. 

Hence, in this paper, we present several contributions
\begin{itemize}
    \item We propose a novel simulator for aerial transportation and manipulation with quadrotor MAVs equipped with passive mechanisms. More specifically, we provide a set of models as well as planning and control algorithms. These are directly accessible from a flexible interface that the user can leverage to define the connection mechanisms, payload, and MAV types.  
    \item We model the full hybrid system dynamics with slack and taut suspended cables including the collision model for the transient dynamics between the two different system dynamics. \textbf{To the best of our knowledge, this is the first time that a full collision analytical model, with a provable closed-form analytical solution, between multiple quadrotors and payload via cables has been developed.} It enables us to provide the complete dynamics of aerial transportation and manipulation with MAVs using passive mechanisms like cables and rigid links.
    \item A set of comprehensive simulation and real-world experiments to validate the fidelity and accuracy of the modules and algorithms including the proposed collision model for multiple quadrotors carrying a rigid-body payload. To the best of our knowledge, this is also the first time real-world experiments are carried out to examine and validate the collision model for multiple quadrotors carrying a rigid-body payload.
\end{itemize}
\textbf{In Fig.~\ref{fig:sample-system}, we show some of the sample systems and the interface that the user can use to define the system setup, controller, and planner they would like to test in RotorTM. Moreover, we release the code of the simulator in MATLAB as well as in Python/ROS, to the community to accelerate the research in aerial transportation and manipulation. We believe that this is the first time the research community gets access to an open-source comprehensive simulation framework for modeling, control, and planning with micro aerial robots transporting or manipulating payloads via passive attach mechanisms.}

We organize the paper as follows. Section~\ref{sec:related-works} offers an overview of existing approaches and corresponding simulation tools for aerial transportation and manipulation. Section~\ref{sec:overview} provides the overview of our simulator features. In Section~\ref{sec:modeling}, we discuss the system dynamics for aerial transportation and manipulation with aerial robots using passive mechanisms. Section~\ref{sec:hybrid-model} introduces the hybrid system model when employing suspended cables as transportation and manipulation mechanism. We provide full collision models of the hybrid system dynamics. Section~\ref{sec:experimental_results} shows simulation and experimental results. These show the simulator's fidelity compared to real-world settings and its benefit in facilitating a rapid transition to real-world deployment. Section~\ref{sec:conclusion} concludes the work and proposes multiple future research directions.

\section{Related Works}\label{sec:related-works}
\subsection{Hybrid Cable Dynamics Modeling}
It is challenging to simulate cables compared to other passive mechanisms, especially the hybrid dynamics cases introduced by the suspended cables. When the cables switch from slack to taut condition, the payload's and robots' velocity transitions to new values due to the collision along the cable direction. Many researchers simplify this problem by assuming the cables are always taut. For example, based on this assumption,  works like~\cite{wu2014geometric, geng2022jais, pratik2020icra} present planning algorithms and nonlinear controllers on multi-MAVs transporting a rigid-body payload validated through a customized simulator. However, it is essential to consider this condition, especially when external disturbances like wind or human interaction exist, as the cable could transition from being taut to slack and vice versa.

It is typical in the research literature to model the cable as rigid links or elastic springs, with a joint at both ends of the cable connecting the robots and the payload. For example, in~\cite{klausen2015icuas}, the authors model the cable as a rigid link and develop a MATLAB-based numerical simulator to test their proposed nonlinear controller on the system of using a single MAV transporting a suspended payload via a cable. 

%The main drawback of these approaches is considering cables as rigid links or elastic springs. However, these assumptions do not hold, especially for cables with small cross-section areas that are made with materials like nylon. More specifically, cables will not generate axial pushing forces to the payload or the quadrotor, while a rigid link or elastic spring modeled in the aforementioned simulators do. 

Another way to circumvent modeling the hybrid dynamics introduced by suspended cable is to model the cable as a series of rigid links with joints between every pair of links like in~\cite{kotaru2020hose}. The main drawback of this approach is that it must include and connect a substantial number of links in series to simulate a cable with sufficient fidelity. It will immediately introduce a high computational burden to the simulator preventing its scalability to multiple robots and long cables. 

Other works instead directly model the hybrid dynamics. For instance, \cite{sreenath2013cdc, RSS2013Koushil, tang2015icra} suggest modeling collision between quadrotors and the suspended payloads as perfectly inelastic collision. \textbf{However, in these three works there is NO explicit mathematical modeling of how the system state changes when the collision occurs as the cables transition from slack to taut. Hence, the models in \cite{sreenath2013cdc, RSS2013Koushil, tang2015icra} cannot be implemented in simulation to simulate collision, and also cannot be validated in the real world.} \cite{cruz2017cable} provides complete modeling of the collision between a quadrotor and a point-mass payload. Based on the aforementioned modeling, other works~\cite{cruz2015acc,zeng2020ral} propose control and planning algorithms considering the hybrid dynamics and test the algorithms in the simulators they developed. For multi-MAVs case, in~\cite{Bisgaard2009Modeling}, a generic collision model between one helicopter and one rigid body payload is derived and examined in numerical simulation. These results show that the collision model between multiple quadrotors and a payload has not been investigated yet, and no simulator incorporates the collision between quadrotors and payloads through cables. Moreover, none of the previously mentioned works~\cite{RSS2013Koushil, Bisgaard2009Modeling} examines their proposed collision model with real-world experiments. 

\subsection{Simulation Tools}
Simulation is a valuable tool for researchers to collect data, test algorithms, and prototype systems. It can help researchers to verify and validate their algorithms before deploying them in the real-world system~\cite{Lee2018,kotaru2017acc}. It can also provide easy access to gather data to train and test learning algorithms when considering recent research on novel machine learning methods like reinforcement learning or meta-learning on a cable-suspended payload with MAVs\cite{Belkhale2021metalearning,faust2017rl,hua2021tmorl}. However, popular open-source MAV simulators like~\cite{airsim,song2020flightmare} can simulate only MAVs without any mechanisms like cables or rigid links for physical environment interaction. 

Other researchers utilize existing simulators like Gazebo with ODE~\cite{russell2008ode} as its physics engine. For instance, in~\cite{okoli2019report}, the authors use rigid links to represent cables in Gazebo for a cable-drive parallel robot. However, ODE and some other state of the art of physics simulation engines, including PyBullet~\cite{coumans2019pybullet}, DART~\cite{lee2018pydart}, only consider rigid body collision, meaning they can simulate collision when 2 rigid bodies collide with each other. There is only one way to simulate multi-robot transportation with cables in these physics engines, which is treating cables as rigid links so that the previously mentioned physics engine can detect the collision between the cable and the quadrotor or between the cable and the payload.

However, a cable is NOT a rigid link. For two objects attached at the two ends of a cable, if the relative velocity makes them move towards each other when the cable is taut, there will be no collision because this would make the cable slack. However, it would never be possible to simulate such an effect if a cable is considered a rigid link or an elastic spring. Hence, there is not a fair comparison between our simulator and the previously mentioned physics engines. The aforementioned engines cannot simulate some important physical phenomena that we report in this work. 

Some researchers propose their own customized simulators for their specific research purposes. For example, the author in~\cite{Lee2018} proposes a nonlinear controller for multi-MAVs transporting a rigid-body payload, assuming cables are rigid links, and examines the controller in a self-developed numerical simulation.  In~\cite{hossein2018icuas, rastgoftar2018calm, kotaru2017acc}, the authors instead model the cables as elastic springs. They design corresponding controllers based on this assumption for multi-MAVs and single MAV transporting a payload via cables and validate them in self-developed simulators. However, none of the aforementioned simulators open-sourced the code. 

%Similar efforts can also be found in research using Gazebo~\cite{koenig2004gazebo} to simulate cable-drive parallel robots.

Researchers also present several open source numerical simulation tools for  specific research purposes related to aerial transportation and manipulation. For example, in~\cite{wang2018icra}, the authors show a decentralized control framework for multiple quadrotors transporting a payload via rigid links. They evaluate the controller with their self-developed Python-based simulator and visualize the results in Rviz. The authors in~\cite{tagliabue2019ijrr} use high-fidelity simulator RotorS~\cite{Furrer2016} to simulate a team of MAVs transporting a rigid body payload via spherical ball joints. Both of the simulators mentioned above are open source. However, this class of simulators is not suited to simulate aerial vehicles endowed with cables and the corresponding hybrid dynamics.

The aforementioned modeling gaps as well as the limitations of existing simulators clearly show the need to design, as proposed in this work, a complete open-source solution that analytically mimics the behavior of cable mechanisms including hybrid dynamics for multiple physically interconnected vehicles. This will allow researchers to have easy access to a simulator tool for designing appropriate control and planning strategies. Therefore, our simulator also includes an inelastic collision model for aerial transportation and manipulation with one or multiple quadrotors via suspended cables. \textbf{In our model, we explicitly model the system state transition, i.e, model how the payload’s linear and angular velocity and quadrotors’ velocity change after the collision happens. The model derivations are shown in Eqs.~(\ref{eq:modelHybridmultiple})-(\ref{eq:multi-equal-attach-vel}), (\ref{eq:multi-after-collision-solution})-(\ref{eq:multi-after-collision-solution-final})}. Furthermore, we implement this model in our simulator and validate it with real-world experiment results.
\begin{table}[t]
\caption {Notation table\label{tab:notation}} 
\centering
%\newcolumntype{s}{}
\begin{tabularx}{0.48\textwidth}{>{\hsize=0.58\hsize}X >{\hsize=1.42\hsize}X}
    \hline\hline
 $\worldf$, $\loadf$, $\robotf{k}$ & inertial frame, payload frame, $k^{th}$ robot frame\\
 $\loadmass,m_k\in\realnum{}$ & mass of payload, $k^{th}$ robot\\
 $\loadpos,\loadvel,\loadacc\in\realnum{3}$ & payload's position, velocity, acceleration in $\worldf$\\
$\robotpos{k}, \robotvel{k}, \robotacc{k}\in\realnum{3}$ &$k^{th}$ robot's position, velocity, acceleration in $\worldf$\\
 $\robotrot{k}, \loadrot\in\SOthree$& $k^{th}$ robot's, payload's attitude with respect to $\worldf$ \\
 $\vecq_{k},\vecq_{L}$& quaternion representation of $\robotrot{k}, \loadrot$\\
 %$\robotrot{} \in\SOthree$&orientation of robot with respect to $\worldf$ \\
 $\loadangvel$, $\loadangacc\in\realnum{3}$ & angular velocity, acceleration of payload in $\loadf$\\
  $\robotangvel{k}\in\realnum{3}$& angular velocity of $k^{th}$ robot in $\robotf{k}$\\
  $f_{k}\in\realnum{}$, $\matM_{k}\in\realnum{3}$&total thrust, moment at $k^{th}$ robot in $\robotf{k}$.\\
 $\inertiaload,\inertia_{k}\in\realnum{3\times3}$   &  moment of inertia of payload, $k^{th}$ robot\\%\hline
 $\cablevec{k}\in S^2$&unit vector from $k^{th}$ robot to attach point in $\worldf$\\
 $l_{k}\in\realnum{}$ & cable length of the $k^{th}$ cable\\
 $\vecrho{k}\in\realnum{3}$&position of $k^{th}$ attach point in $\loadf$\\
    \hline\hline
\end{tabularx}
\end{table}

\begin{table*}[!t]
\caption {Software Module Interface\label{tab:software-interface}} 
\centering
%\newcolumntype{s}{}
\begin{tabularx}{\textwidth}{>{\hsize=0.1\hsize}X|>{\hsize=0.45\hsize}X|>{\hsize=0.65\hsize}X|>{\hsize=0.5\hsize}X |>{\hsize=0.15\hsize}X |>{\hsize=0.15\hsize}X}
    \toprule\hline\hline
 Module    & Options & Input & Output & Equations & Model \\[0.25em]\hline
 \multirow{2}{4em}{PlTraj} & Circle & t, Duration, Radius &\multirow{2}{4em}{$\robotpos{L,des}, \robotvel{L,des}, \robotacc{L,des},$\\ $\vecq_{L,des},\robotangvel{L,des}$} &\hspace{1em}\rule[0.5ex]{2em}{0.55pt} &\multirow{2}{4em}{\hspace{1em}\rule[2ex]{4em}{0.55pt}} \\\cline{2-3}\cline{5-5} 
& Minimum\_kth\_derivative  & t & & \eqref{eqn:poly_spline}$-$\eqref{eqn:cost_integral}\\\hline  \multirow{2}{4em}{QdTraj} & Circle  & t, Duration, Radius   &\multirow{2}{4em}{$\robotpos{des}, \robotvel{des}, \robotacc{des}$,\\ $\psi_{des},\dot{\psi}_{des}$}&\hspace{1em}\rule[0.5ex]{2em}{0.55pt} &\multirow{2}{4em}{\hspace{1em}\rule[2ex]{4em}{0.55pt}} \\\cline{2-3} \cline{5-5} 
           & Minimum\_kth\_derivative  & t   & &\eqref{eqn:poly_spline}$-$\eqref{eqn:cost_integral}\\ \hline
 \multirow{4}{4em}{PlCtrl} & Single\_cable\_geometric\_control &$\robotpos{L,des}, \robotvel{L,des}, \robotacc{L,des},\psi_{des},\dot{\psi}_{des},$ & $f,\matM$ & \eqref{eq:control-single-cable-plpos-control}$-$\eqref{eq:control-single-cable-qdthrust-control} & \eqref{eq:single-kinematics}$-$\eqref{eq:single-slack-robot-rotation-dyn} \\\cline{2-6}
   & \multirow{2}{4em}{Multi\_cable\_geometric\_control} & $\robotpos{L,des}, \robotvel{L,des}, \robotacc{L,des},\vecq_{L,des},\robotangvel{L,des},$  & \multirow{2}{4em}{$f_1,\matM_1,\cdots,f_n,\matM_n$} &\multirow{2}{4em}{\eqref{eq:control-multi-cable-pl-control}$-$\eqref{eq:control-multi-cable-qd-control}} & \multirow{2}{4em}{\eqref{eq:multi-payload-eqn-motion-translation}$-$\eqref{eq:multi-slack-robot-dyn}}\\ 
 & &  $\psi_{1,des},\cdots,\psi_{n,des}, \dot{\psi}_{1,des},\cdots,\dot{\psi}_{n,des}$  & & & \\\cline{2-6}
           & Multi\_rigid\_links\_control  & $\robotpos{L,des}, \robotvel{L,des}, \robotacc{L,des},\vecq_{L,des},\robotangvel{L,des},$  & $f_1,\matM_1,\cdots,f_n,\matM_n$ &\eqref{eq:control-structure-control}&\eqref{eq:structure-translation-dyn}$-$\eqref{eq:structure-thrust-moment-mapping}\\ \hline
 \multirow{2}{4em}{QdCtrl}& Geometric\_control  &  $\robotpos{k,des}, \robotvel{k,des}, \robotacc{k,des}, \psi_{k,des},\dot{\psi}_{k,des}$  &$f,\matM$& \multirow{2}{4em}{\hspace{1em}\rule[2ex]{4em}{0.55pt}}&\multirow{2}{4em}{\hspace{1em}\rule[2ex]{4em}{0.55pt}}\\\cline{2-4} 
           & Quadrotor\_attitude\_control  & $\vecq_{des},\robotangvel{des}$  & $\matM$ & & \\
    \hline\hline
\end{tabularx}
\end{table*}
\begin{figure*}
    \centering
    \includegraphics[width=\textwidth]{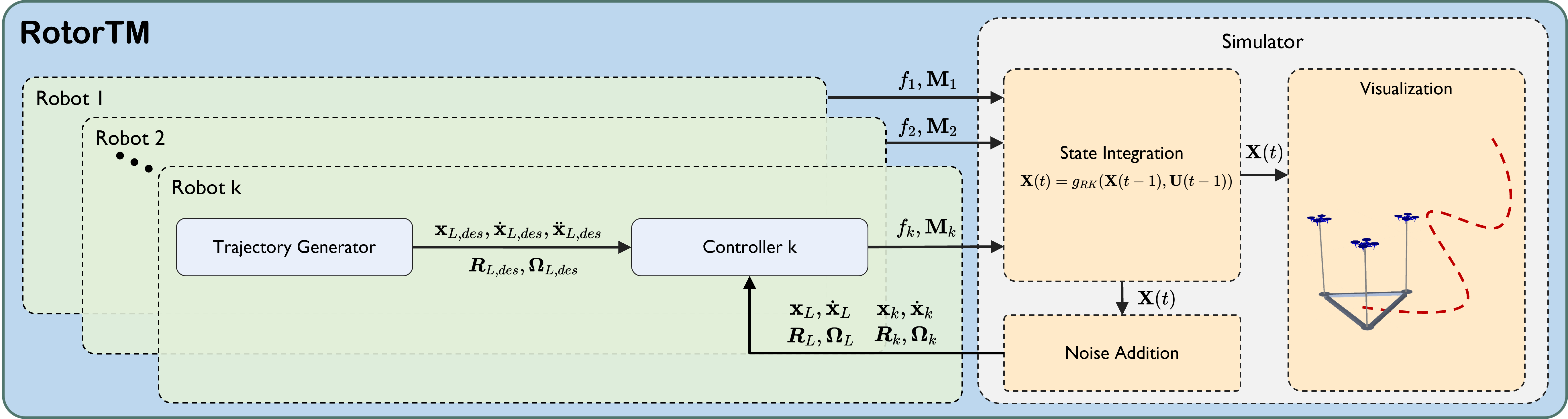}
    \caption{RotorTM software modules overview. }
    \label{fig:control_block_diagram}
\end{figure*}
\section{Overview}\label{sec:overview}

We consider several systems with passive connection mechanisms: i) a point-mass payload suspended from a single quadrotor, ii) a rigid-body payload carried by $n$ quadrotors with either rigid links or cables. The simulator can be easily generalized by the user, who can customize different systems in terms of the number and types of vehicles and payloads according to the specific user's needs. 

We summarize the relevant variables in our setup in Table~\ref{tab:notation}, as well as the possible options and interface of the software mentioned above modules in Table~\ref{tab:software-interface}. We use $\matX(t)$ to denote the state vector of the entire system and $\matU(t)$ to represent the input vector of the entire system at the time $t$. The simulator will start the system evolution with the initial state $\matX(0)$. Based on the user's choice of the trajectory generator, controller, and system in the user interface, the simulator will integrate the states and visualize the system accordingly.

\subsection{System States}
\subsubsection{Single Quadrotor}
In this case, since we only consider one quadrotor, we will drop the subscript of the robot's state for notational convenience shown in Table~\ref{tab:notation}. For the payload suspended from a quadrotor system through a cable, the system state and input vector are  
\begin{equation}
    \matX = \begin{bmatrix}\loadpos^{\top}, \loadvel^{\top}, \robotpos{}^{\top}, \robotvel{}^{\top}, \vecq_{}^{\top},   \robotangvel{}^{\top}\end{bmatrix}^{\top},\matU =  \begin{bmatrix}f,\matM^{\top}\end{bmatrix}^{\top}.\label{eq:single-state-vector}
\end{equation}

\subsubsection{Multiple Quadrotors Via Cables}
For a system with a rigid body payload suspended from $n, n\geq2,$ quadrotors, the state vector $\matX$ of the system is 
\begin{equation}
    \matX = \begin{bmatrix}\matX_{L}^{\top}, \matX_{1}^{\top},\cdots,\matX_{k}^{\top},\cdots,\matX_{n}^{\top}\end{bmatrix}^{\top},\label{eq:multiple-state-vector}
\end{equation}
where $\matX_{L}\in\realnum{13}$ represents the payload's state and $\matX_{k}\in\realnum{13}$ represents the $k^{th}$ quadrotor's state, $k = 1,\cdots,n$. $\matX_{L}$ and $\matX_k$ are defined as 
\begin{equation*}
    \matX_{L} = \begin{bmatrix}\loadpos^{\top}, \loadvel^{\top}, \vecq_{L}^{\top},\loadangvel^{\top}\end{bmatrix}^{\top},~\matX_{k} = \begin{bmatrix}\robotpos{k}^{\top}, \robotvel{k}^{\top}, \vecq_{k}^{\top},\robotangvel{k}^{\top}\end{bmatrix}^{\top}.
\end{equation*}
%where $\vecq_{L},\vecq_{k}$ is the quaternion representation of the payload's and $k^{th}$ robot's orientation with respect to the world frame $\worldf$. 
The system control input is
\begin{equation}
    \matU =  \begin{bmatrix}f_1,\matM_1^{\top},\cdots,f_k,\matM_k^{\top},\cdots,f_n,\matM_n^{\top}\end{bmatrix}^{\top}.\label{eq:multiple-input-vector-cable}
    %\matU =  \begin{bmatrix}\matU_1^{\top},\cdots,\matU_k^{\top},\cdots,\matU_n^{\top}\end{bmatrix}^{\top},\label{eq:multiple-input-vector}
\end{equation}

%\begin{figure}[t]
%    \centering
%    \includegraphics[width=\columnwidth]{figs/multi-robot-rigid-links.png}
%    \caption{Multiple quadrotors manipulating a rigid-body payload through rigid links.}
%    \label{fig:multi-robot-rigid}
%\end{figure}

\subsubsection{Multiple Quadrotors Via Rigid Links}
We can model the system as a unique rigid body if $n$ quadrotors connect a rigid body payload via rigid links. Hence we only need to simulate the position $\robotpos{c}$, velocity $\robotvel{c}$ of the center of mass of the entire structure, and the orientation $\vecq_c$, angular velocity $\robotangvel{c}$ of the entire structure. Hence, the simulated state vector $\matX$ is 
\begin{equation}
    \matX = \begin{bmatrix}
     \vecx_c^{\top},\dot{\vecx}_c^{\top},\vecq_c^{\top},\bm{\Omega}_c^{\top}
    \end{bmatrix}^{\top}.
\end{equation}
Further, the states of the payload and the quadrotors can be obtained by using the geometric constraints between them. The input vector $\matU$ is the same as Eq.~(\ref{eq:multiple-input-vector-cable}).
%\begin{equation}
%    \matU =  \begin{bmatrix}f_1,\matM_1^{\top},\cdots,f_k,\matM_k^{\top},\cdots,f_n,\matM_n^{\top}\end{bmatrix}^{\top}\label{eq:multiple-input-vector-rigid}.
    %\matU =  \begin{bmatrix}\matU_1^{\top},\cdots,\matU_k^{\top},\cdots,\matU_n^{\top}\end{bmatrix}^{\top},\label{eq:multiple-input-vector}
%\end{equation}

\subsection{Software Modules}
As shown in Fig.~\ref{fig:control_block_diagram}, the simulator mainly consists of five modules, i.e., trajectory generator, controller, system dynamics integration, visualization, and intuitive user interfaces. We release two versions of the simulator. The first one is written in Python and ROS for robotics and control research purposes. The second one is implemented in Matlab with visualization in ROS, tailored for general academic purposes.

\subsubsection{System Dynamics}
We can express the system dynamics modeling in Section~\ref{sec:modeling} for all the possible systems in a general ordinary differential equation as following 
\begin{equation}
    \dot{\matX} = g\prths{\matX,\matU}.
\end{equation}
In the simulator implementation, we leverage Runge-Kutta methods to numerically integrate the above ordinary differential equation from states $\matX\prths{t-1}$ and inputs $\matU\prths{t-1}$ at time $t-1$ to the states $\matX\prths{t}$ at time $t$ as following:
\begin{equation}
    \matX\prths{t} = g_{RK}\prths{\matX\prths{t-1},\matU\prths{t-1}}.
\end{equation}

\subsubsection{Trajectory Generator}
The trajectory generator generates the desired trajectory for the payload or the quadrotor based on the users' preference to directly control the
load or control the quadrotor to track a series of waypoints. We provide several trajectory options in this simulator, including a circular trajectory and minimum-$k^{th}$-derivative trajectory among waypoints in Section~\ref{sec:trajectory_generation}. Therefore, the user can also use this simulator to develop and test motion planning and trajectory generator algorithms. 

\subsubsection{Controller}
The controller takes the desired value for the trajectory generation function and the current states as inputs and computes quadrotors' actions $\matU$ in Eqs.~\eqref{eq:single-state-vector} and \eqref{eq:multiple-input-vector-cable}. The user can develop and test as well customized controllers. The controller provided in the RotorTM is presented in Section~\ref{sec:controller}.

%\begin{algorithm}[t]
%\caption{Given the initial state $\matX_0$, trajectory generator and controller, simulate the motion of the payload and the quadrotor team from 0 to T.}\label{alg:sim}
%\begin{algorithmic}
%\Function{Main}{} 
%\State $\matX \gets \matX_0$
%\State $t \gets 0$
%\State QdTraj([], $\matX_0$, $\matX_f$)
%\State PlTraj([], $\matX_0$, $matX_f$) 
%\State Dyn([], [], params, QdTraj, PlTraj, PlCtrl, QdCtrl) 
%\For{$iter = 1:max\_iter$}
%    \If{Hybrid}
%      \State $CollideFlag$ = CheckCollisionFlag$\prths{\matX}$
%        \If{CollideFlag}
%          \State $\matX$ = ResetState($\matX$, params)
%        \EndIf
%    \State $StopCond$ = SetStopCondition$\prths{t,\matX}$
%    \EndIf
%    \State $\matX =$ RK45$\prths{\text{Dyn}\prths{t,\matX},StopCond}$
%    \State Visualize$\prths{\matX}$
%\EndFor
%\EndFunction
%\end{algorithmic}
%\end{algorithm}

\section{System Dynamics}\label{sec:modeling}
In this section, we introduce the dynamics models in the simulator. The models are developed based on the following assumptions: 
\begin{enumerate}
    \item The drag on the payload and quadrotor is negligible;
    \item The rotor dynamics time evolution is much faster compared to the high-level position and attitude controller;
    \item The massless cable connects to the robot's center of mass;
    \item The aerodynamic effect among the robots and the payload is neglected.
\end{enumerate}
The above assumptions are reasonable since they have been successfully applied in several works~\cite{LoiannoRAL2018, guanrui2021ral, tang2015icra} and tested in real-world settings at sustained speed. In addition, users can also easily customize the simulator according to their specific needs and contribute to its generalization. In Section~\ref{sec:experimental_results}, we will also show the comparison between the simulation results and real-world experiments proving the efficacy and accuracy of the proposed simulator.

\subsection{Single Quadrotor}\label{sec:single-modeling}
\begin{figure}[!t]
    \centering
    \includegraphics[width=0.6\columnwidth]{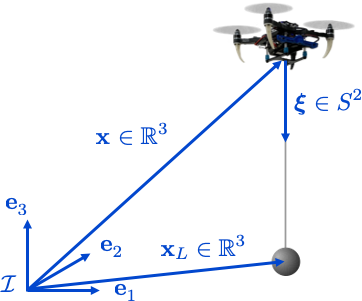}
    \caption{Depiction of Single MAV transporting a payload via a suspended cable.}~\label{fig:single-robot}
\end{figure}
We introduce the dynamics of a quadrotor connected to a point mass payload through a massless cable, as shown in Fig.~\ref{fig:single-robot}. The system is a hybrid system since its dynamics differ depending on whether the cable is taut or slack.
\subsubsection{Taut Cable}\label{sec:single-modeling:taut}
The system configuration space here is $SE(3)\times S^2$. As shown in Fig.~\ref{fig:single-robot}, the geometric constraint between the quadrotor and the payload is
\begin{equation} 
    \loadpos = \robotpos{} + l\cablevec{},\label{eq:geometric-pos-robot-ptmass}
\end{equation}
where $\cablevec{}\in S^2$ is a unit vector from the robot's center of mass to the payload. Based on \Eqref{eq:geometric-pos-robot-ptmass}, we can have 
\begin{equation} 
   \cablevec{} = \frac{\loadpos - \robotpos{}}{l}, ~ \cabledotvec{} =\frac{\loadvel - \robotvel{}}{l}, ~\cableddotvec{} = \frac{\loadacc - \robotacc{}}{l}.\label{eq:geometric-robot-ptmass}
\end{equation}
According to~\cite{sreenath2013cdc}, by applying  the  Lagrange--d’Alembert  principle, we obtain the system's equations of motion
\begin{equation}
\frac{d\loadpos}{dt} = \loadvel, ~\frac{d\robotpos{}}{dt} = \robotvel{}, ~\dot{\vecq}_{} = \half\hat{\robotangvel{}}_{}\cdot\vecq_{},\label{eq:single-kinematics}
\end{equation}
\begin{equation}
\prths{m+\loadmass}\prths{\loadacc + \vecg} = \prths{\cablevec{}\cdot f\robotrot{}\axis{3}{}-m l\prths{\cabledotvec{}\cdot\cabledotvec{}}}\cablevec{},\label{eq:single-load-lagrange-eom}
\end{equation}
\begin{align}
m l\prths{\cableddotvec{}+\prths{\cabledotvec{}\cdot\cabledotvec{}}\cablevec{}} & = \cablevec{}\times\prths{\cablevec{}\times f\robotrot{}\axis{3}{}},\label{eq:single-quad-lagrange-eom}\\
\matM &= \inertia\robotangacc{} + \robotangvel{}\times\inertia\robotangvel{},\label{eq:single-robot-rotation-dyn}
\end{align}
where $\vecg = g\axis{3}{}$, $g = 9.81 \si{m/s^2}$ and $\axis{3}{} = \left[0~0~1\right]^{\top}$, and $\hat{\robotangvel{}}$ is the skew-symmetric matrix of the quadrotor angular velocity $\robotangvel{}$. 
By using Eqs.~\eqref{eq:geometric-robot-ptmass}-\eqref{eq:single-robot-rotation-dyn},  
we can write the system's equations of motion written in the standard form
\begin{equation}
\dot{\matX} = g_p\prths{\matX,\matU},
\label{eq:matrixform}
\end{equation}
where $g_p$ corresponds to the dynamics function when the cable is taut, $\dot{\matX} = \begin{bmatrix}\loadvel^{\top}, \loadacc^{\top}, \robotvel{}^{\top}, \robotacc{}^{\top}, \dot{\vecq}_{}^{\top}, \robotangacc{}^{\top}\end{bmatrix}^{\top}$.
% ------------------------------------------------------------------------------- %

\subsubsection{Slack Cable}\label{sec:single-modeling:slack}
In this case, the tension in the cable is zero and the system configuration space is $SE(3)\times\realnum{3}$. Moreover, the quadrotor and the payload evolve as two independent systems. Therefore, the equations of motion are
\begin{align}
\loadmass\prths{\loadacc + \vecg} &= 0,~f\robotrot{}\axis{3}{} = m\prths{\robotacc{} + \vecg},\label{eq:single-slack-dyn}\\
 \matM &=\inertia\angacc + \angvel\times\inertia\angvel \label{eq:single-slack-robot-rotation-dyn}.
\end{align}
Eqs.~\eqref{eq:single-slack-dyn} and \eqref{eq:single-slack-robot-rotation-dyn} can be written similarly to \Eqref{eq:matrixform}.
%in a similar way we have in the Section.~\ref{sec:single-modeling:taut}:
%\begin{equation}
%\dot{\matX} = g_z\prths{\matX} + h_z\prths{\matX}\matU,
%\end{equation}

\subsection{Multiple Quadrotors}\label{sec:multi-modeling}
\subsubsection{Cable Mechanism}
%        \subfigure[Multiple quadrotors transporting a rigid-body mass payload via rigid links.\label{fig:multi-robot-rigid}]{
%    \includegraphics[width=0.5\textwidth]{figs/multi-robot-rigid-links.png}}
\begin{figure}[!t]
    \centering
    \includegraphics[width=\columnwidth]{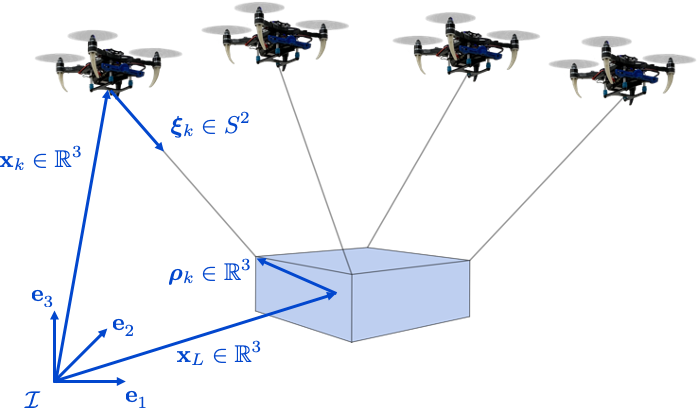}
    \caption{Depiction of multiple MAV transporting a rigid-body payload via suspended cables.}~\label{fig:multi-robot-cable}
\end{figure}
In this section, we model the dynamics where $n$ quadrotors cooperatively transport a rigid body payload using massless cables with respect to $\worldf$ shown in Fig.~\ref{fig:multi-robot-cable}. The $k^{th}$ quadrotor is connected to the payload at an attach point, whose position is $\mathbf{p}_k\in\realnum{3}$ with respect to $\worldf$. As the payload is modeled as a rigid body, we have
\begin{equation}
    \mathbf{p}_k = \loadpos + \loadrot\vecrho{k}\label{eq:attach-rigidbody-payload-geometry}
\end{equation}

Similar to Section~\ref{sec:single-modeling}, we consider the system dynamics with cables in slack and taut conditions. However, in this case, it is more complicated with respect to the single robot case because multiple cables can be in slack conditions simultaneously. Hence, we assume that $n_p$ quadrotors' cables are taut and $n_z$ quadrotors' cables are slack. Then we have $n = n_p + n_z$.
\begin{figure}[!t]
    \centering
    \includegraphics[width=\columnwidth]{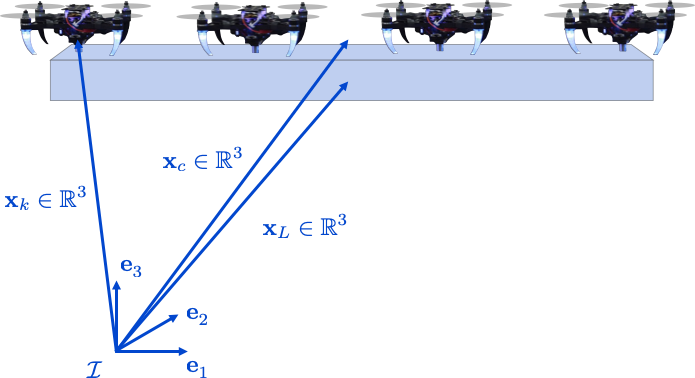}
    \caption{Depiction of multiple MAVs transporting a rigid-body payload via rigid links.}~\label{fig:multi-robot-rigid}
\end{figure}
Hence, the system configuration space is $\prths{SE(3)}^{n_z + 1}\times\prths{S^2\times SO(3)}^{n_p}$. Since only the taut cables will generate tension forces to affect the payload's motion, the payload's equations of motion are related to the inputs introduced by these quadrotors. Without loss of generality, let $\matI_p = \crbrc{1,\cdots,n_p}$ be the set of indices of the taut cables and the corresponding quadrotors, whereas $\matI_z =\crbrc{n_p+1,\cdots, n}$ be the set of indices of the slack ones. The quadrotors with taut cables have the following geometric relationship with the payload's center of mass as depicted in Fig.~\ref{fig:multi-robot-cable}
\begin{equation}
\begin{split}
    \robotpos{k} &= \mathbf{p}_k - l_{k}\cablevec{k} \\
    &= \loadpos+\loadrot\vecrho{k} - l_{k}\cablevec{k},~ k \in \matI_p.\label{eq:multi-geometric-relationship}
\end{split}
\end{equation}
By utilizing the Lagrange--d'Alembert principle~\cite{Leecdc2014,wu2014geometric}, the payload's equations of motion are
\begin{equation}
 \sum_{k=1}^{n_p} \bar{\vecu}_{k}^{||} = \loadmass\prths{\loadacc + \vecg}\label{eq:multi-payload-eqn-motion-translation}
\end{equation}
\begin{equation}
  \sum_{k=1}^{n_p} \hatvecrho{k}\matR_{L}^{\top}\bar{\vecu}_{k}^{||} =\inertiaload\loadangacc+\hat{\angvel}_L\inertiaload\loadangvel,\label{eq:multi-payload-eqn-motion-rotation}
\end{equation}
where 
 \begin{align*}
 \bar{\vecu}_{k}^{||} &= \vecu_{k}^{||} - m_k l_{k}\twonorm{\omega_{k}}^2\cablevec{k}  - m_k\cablevec{k}\cablevec{k}^{\top}\veca_{k},\\
 \veca_{k}&=\loadacc + \vecg-\loadrot\hatvecrho{k}\loadangacc+\loadrot\hat{\angvel}_L^{2}\vecrho{k},\\
\inputforce{k} &= f_k\robotrot{k}\axis{3}{}, ~\inputpara{k} = \cablevec{k}\cablevec{k}^{\top}\inputforce{k},~
\inputperp{k}  = \prths{\matI_{3\times3} - \cablevec{k}\cablevec{k}^{\top}}\inputforce{k}.
 \end{align*}
In the above equations, $\bar{\vecu}_{k}^{||}$ represents the effective force acting on the payload from the $k^{th}$ quadrotor, and $\vecu_{k}^{||}$ is the $k^{th}$ quadrotor's total rotors force $\vecu_{k}$ projected along the cable direction. In addition, $\veca_{k}$ denotes the acceleration with respect to $\worldf$ at the $k^{th}$ attach point.  

Besides the payload dynamics mentioned above, we can obtain the quadrotors' equations of motion by using the same principle. All the quadrotors have the same equations of motion for rotational motion similar to \Eqref{eq:single-robot-rotation-dyn}. However, for translation, quadrotors with taut cables differ from those with slack cables since they are constrained to be on a sphere centered at each corresponding attach point. Therefore, in the case of the taut cables, we have
 \begin{equation}
\hat{\cablevec{}}_{k}^2\prths{\inputforce{k}-m_k\veca_k} = m_k l_{k}\prths{\cableddotvec{k} +\twonorm{\cabledotvec{k}}^2\cablevec{k}}, ~k\in\matI_p.
\label{eq:multi-taut-robot-dyn}
 \end{equation}
Conversely, for quadrotors with slack cables, we have
 \begin{equation}
     \inputforce{k} = m_k(\robotacc{k} + \vecg), ~k\in\matI_z\label{eq:multi-slack-robot-dyn}.
 \end{equation}
By writing the Eqs.~\eqref{eq:multi-geometric-relationship}-\eqref{eq:multi-slack-robot-dyn} in standard form, we obtain
\begin{equation}
    \dot{\matX} = g_{j}\prths{\matX, \matU}, \, j\in [0,n],
\end{equation}
where $g_j$ corresponds to the system dynamics when $j$ cables are taut with
\begin{equation}
\dot{\matX} = \begin{bmatrix}\dot{\matX}_{L}^{\top}, \dot{\matX}_{1}^{\top},\cdots,\dot{\matX}_{k}^{\top},\cdots,\dot{\matX}_{n}^{\top}\end{bmatrix}^{\top}.
\end{equation}
\subsubsection{Rigid Link Mechanism} 
In the following, we model the dynamics of $n$ quadrotors cooperatively transporting a rigid-body payload using rigid links with respect to $\worldf$ shown in Fig.~\ref{fig:multi-robot-rigid}. Since the quadrotors and the payload are connected via rigid links, we can model them as a single rigid-body structure with inputs given by the thrust and moment generated by each robot. Hence, the configuration space is $SE(3)$. As mentioned in Section~\ref{sec:overview}, we simulate the motion of the entire structure and obtain the states of the payload and the quadrotors based on the geometric constraints imposed by the rigid links
\begin{align}
\frac{d\robotpos{c}}{dt} = \robotvel{c},~m_c\robotacc{c} &= \robotrot{c}f_c\axis{3}{} - m_c\vecg, \label{eq:structure-translation-dyn}\\
\dot{\vecq}_{c} = \half\hat{\robotangvel{}}_{c}\cdot\vecq_{c},~\matM_c &= \inertia\robotangacc{c} + \robotangvel{c}\times\inertia\robotangvel{c},\label{eq:structure-rotation-dyn}
\end{align}
where $f_c\in\realnum{}$ is the sum of all the quadrotors' rotor thrusts, $\matM_c\in\realnum{3}$ is the net moments contributed by all the rotor forces on the payload and $m_c$ is the total mass of the entire structure. The mapping from quadrotor rotor thrust and moments to the net thrust $f_c$ and the net moment $\matM_c$ is     
\begin{equation}
\begin{split}
    &\begin{bmatrix}
      f_c\\\matM_c
    \end{bmatrix} =\matA\matU = \sumn{k}\matA_k\begin{bmatrix}
      f_k\\\matM_k
    \end{bmatrix}\label{eq:structure-thrust-moment-mapping},%\\
%    \matA_k =& \begin{bmatrix}
%      1&0&0&0\\
%      y_k&\cos{\psi_k}&-\sin{\psi_k}&0\\
%      -x_k&\sin{\psi_k}&\cos{\psi_k}&0\\
%      0&0&0&1\\
%    \end{bmatrix}, 
\end{split}
\end{equation}
where $\matA_k$, defined in~\cite{LoiannoRAL2018}, is the mapping  matrix accounting for the $k^{th}$ quadrotor's thrust $f_k$ and moment $\matM_k$ contributions to the total thrust $f_c$ and moment $\matM_c$. %$\prths{x_k,y_k}$ is the position of the $k^{th}$ quadrotor with respect to the structure's body frame located at the center of mass of the structure, $\psi_k$ is the yaw angle of the $k^{th}$ quadrotor with respect to the structure's frame. 
\begin{figure}[!t]
    \centering
    \includegraphics[width=\columnwidth]{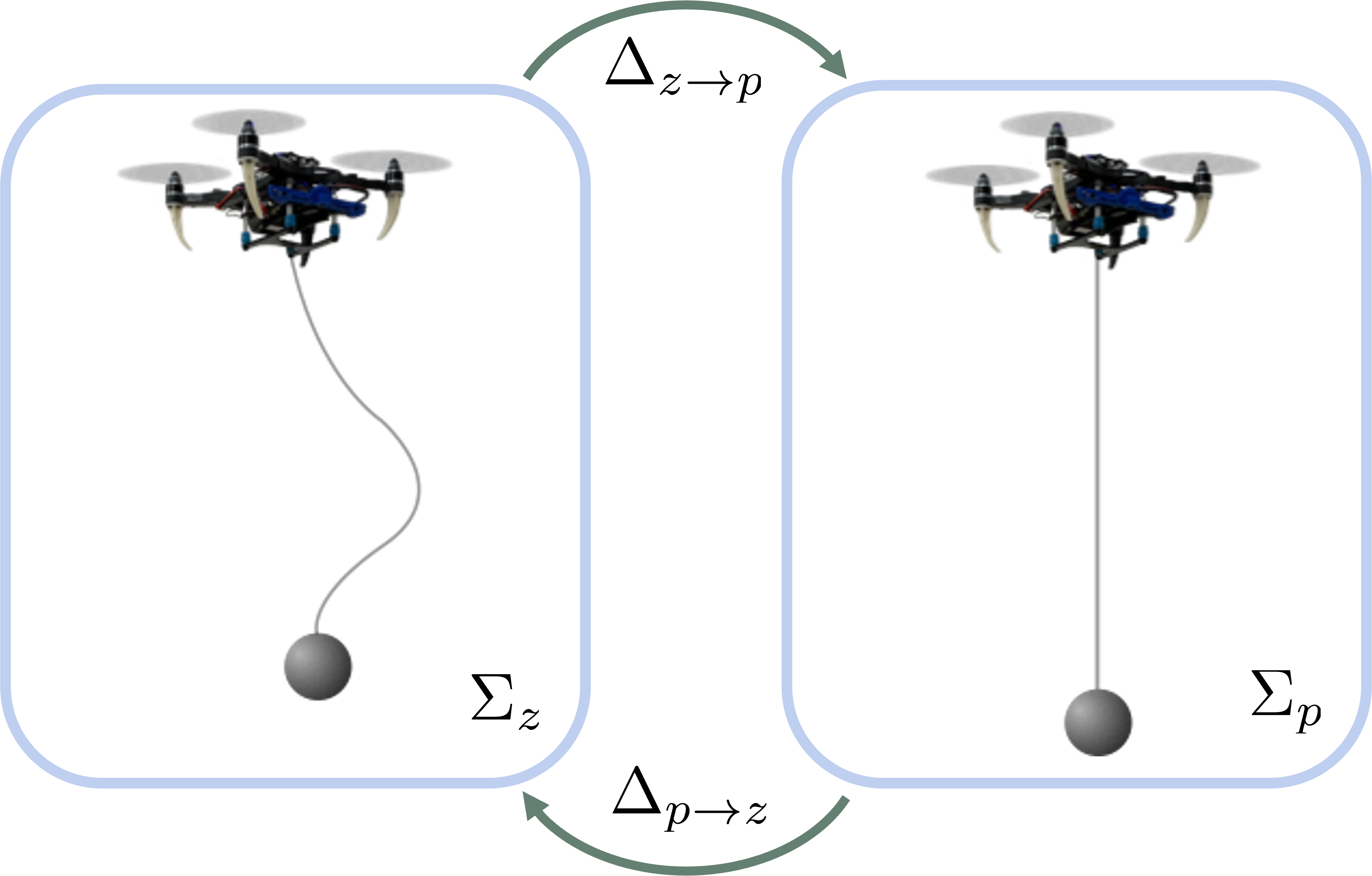}
    \caption{Transition between two systems as the cable in the system becomes slack or reestablishes tension}
    \label{fig:single-hybrid}
\end{figure}
\section{Hybrid System Dynamics}\label{sec:hybrid-model}
Our solution incorporates the challenging case of hybrid dynamics that models cable transitions from being taut to slack and vice-versa (once the tension is reestablished) for one or multiple quadrotors transporting loads via cables. In this section, we model the transition reset maps among different system dynamics considering inextensible cables. 

\begin{figure*}[!t]
    \centering
    \includegraphics[width=\textwidth]{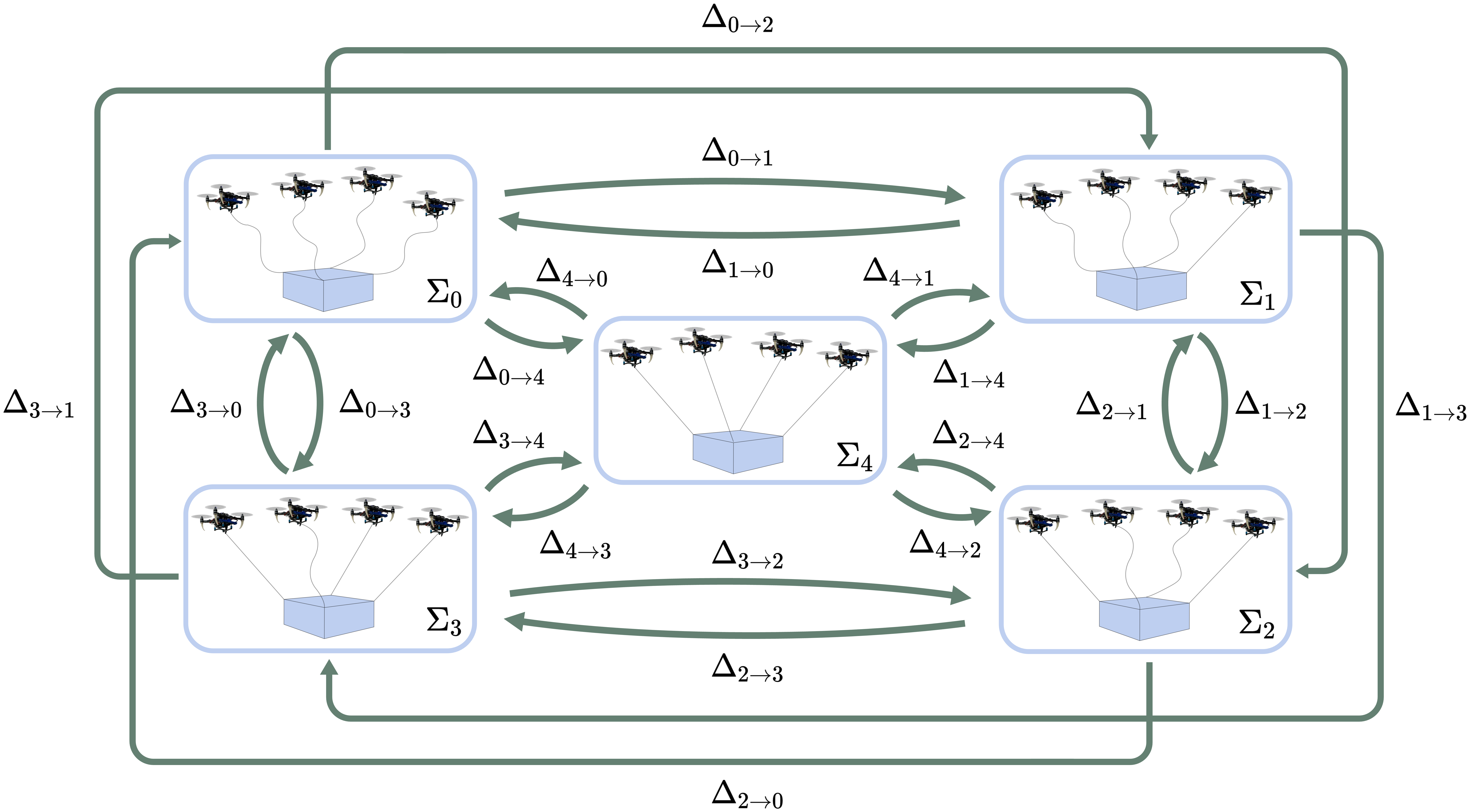}
    \caption{The transition between subsystems when $n=4$ as the cables become slack or reestablish tension based on our collision model developed in Section~\ref{sec:multiple-hybrid-model}. }
    \label{fig:multi-hybrid}
\end{figure*}
\subsection{Single Quadrotor}\label{sec:single-hybrid-model}
A single quadrotor with a cable-suspended payload is a hybrid system because the system dynamics switch between two dynamical models, $\Sigma_p$ and $\Sigma_z$. The hybrid system can be modeled as follows
\begin{equation}
  \text{Taut } \Sigma_p:  \begin{cases}
\dot{\matX} = g_p\prths{\matX,\matU},  & \matX\not\in S_z\vspace{0.5em}\\
\matX^{+} = \Delta_{p\rightarrow z}\prths{\matX^{-}}, & \matX\in S_z%\\
%\matX^{+} = \Delta_{n\rightarrow n}\prths{\matX^{-}}, & \matX\in S_n\\
\end{cases}
\end{equation}
\begin{equation}
  \text{Slack }\Sigma_z:  \begin{cases}
\dot{\matX} =  g_z\prths{\matX,\matU}, & \matX\not\in S_p\vspace{0.5em}\\
\matX^{+} = \Delta_{z\rightarrow p}\prths{\matX^{-}}, & \matX\in S_p
\end{cases}
\vspace{-5pt}
\end{equation}
where $*^{-}$ and $*^{+}$ represent the states before and after reset respectively. We also demonstrate the model in Fig.~\ref{fig:single-hybrid}. Let $d$ denote the distance between the robot and the payload
\begin{equation}
    d = \twonorm{\robotpos{}-\loadpos}.
\end{equation}
Then we can define the guards of the two different system dynamics $S_z$ and $S_p$ as 
\begin{equation}
   \begin{split}
  S_z = \crbrc{\matX~\lvert~ d < l}, \, S_p = \crbrc{\matX~\lvert~ d = l,\dot{d}>0}.
   \end{split} 
\end{equation}
The set $S_z$ represents the system state when the cable becomes slack. When the system state reaches region $S_z$, the system dynamics model will transition from $\Sigma_p$ to $\Sigma_z$ via the transition map $\Delta_{p\rightarrow z}$, which is an identity map. $S_p$ represents the system state when the cable becomes taut. When the system state reaches $S_p$, the system will trigger an inelastic collision between the payload and the robot along the cable direction and transition to $\Sigma_p$ via the reset transition map $\Delta_{z\rightarrow p}$, which will be modeled in the following. 
%Both states in $\Sigma_n$ and $\Sigma_z$ can enter the region $S_n$ and trigger a system state reset $\Delta_{z\rightarrow n}$ and $\Delta_{n\rightarrow n}$ correspondingly.

Let the payload velocity and robot velocity that is orthogonal to the cable direction be $\robotvel{L\perp}$, $\robotvel{\perp}$ and those along the cable be $\robotvel{L||}$ and $\robotvel{||}$ respectively, therefore,
\begin{equation}
\begin{split}
   \loadvel = \robotvel{L\perp} + \robotvel{L||}&, ~\robotvel{} = \robotvel{\perp} + \robotvel{||},\\
   \robotvel{L||} = \cablevec{}\cablevec{}^\top \loadvel &, ~ \robotvel{||}= \cablevec{}\cablevec{}^\top\robotvel{}.\label{eq:single-vel-relationship}
\end{split}
\end{equation}
After the collision, based on the Momentum Conservation Principle, the total system momentum along the cable direction is preserved obtaining
\begin{equation}
m\robotvel{||}^{+} + \loadmass\robotvel{L||}^{+} = m \robotvel{||}^{-} + \loadmass\robotvel{L||}^{-}.\label{eq:robot-load-momentum-conservation}
\end{equation}
Moreover, the collision does not affect the payload's and robot's momentum that is orthogonal to the cable direction. Hence, the payload's and robot's velocity in the orthogonal direction to the cable should remain unchanged. 

Since we model the collision as a perfectly inelastic collision along the cable direction, the projection of the payload's and robot's velocity along the cable should be equal
\begin{equation}
\robotvel{||}^{+} = \robotvel{L||}^{+}, ~\robotvel{L\perp}^{-} = \robotvel{L\perp}^{+}, ~\robotvel{\perp}^{-} = \robotvel{\perp}^{+}. \label{eq:robot-load-inelastic-collision}
\end{equation}
By substituting \Eqref{eq:single-vel-relationship} and \Eqref{eq:robot-load-inelastic-collision} into \Eqref{eq:robot-load-momentum-conservation}, we obtain that
\begin{equation}
\robotvel{||}^{+} = \robotvel{L||}^{+} =\frac{m\cablevec{}\cablevec{}^\top\robotvel{}^{-} + \loadmass\cablevec{}\cablevec{}^\top\loadvel^{-}}{m+\loadmass}.\label{eq:single-vel-aftercollision}
\end{equation}
%Hence with Eqs.~\eqref{eq:single-vel-relationship}-\eqref{eq:single-vel-aftercollision}, we have the reset map $\Delta_{z\rightarrow p}$ as % and $\Delta_{n\rightarrow n}$ as following: 
%\begin{equation}
%\begin{split}
%\robotpos{}^{+} &= \robotpos{}^{-},~\robotpos{L}^{+} = \robotpos{L}^{-},~\vecq_{}^{+} = \vecq_{}^{-},~\robotangvel{}^{+} = \robotangvel{}^{-},\\
%\robotvel{}^{+} &=  \robotvel{Q||}^{+} + \robotvel{Q\perp}^{+} =\robotvel{Q||}^{+} + \robotvel{Q}^{-} - \cablevec{}\cablevec{}^\top\robotvel{Q}^{-},\\
%\robotvel{L}^{+} &=  \robotvel{L||}^{+} + \robotvel{L\perp}^{+} =\robotvel{L||}^{+} + \robotvel{L}^{-} - \cablevec{}\cablevec{}^\top\robotvel{Q}^{-}.\\
%\end{split}
%\end{equation}

\subsection{Multiple Quadrotors}\label{sec:multiple-hybrid-model}
In this section, we model $n$ quadrotors with a suspended rigid-body payload as a hybrid system where the dynamical model switches among $n+1$ different modes, $\Sigma_{j}, j\in[0,n]$. $\Sigma_{j}$ and $\Sigma_{r}$ represent the dynamical model considering $j$ and $r$ taut cables, respectively. Subsequently, we model the entire hybrid system as
\begin{equation}
  \Sigma_{j}:  \begin{cases}
\dot{\matX} = g_{j}\prths{\matX,\matU},  & \matX\not\in S_r\vspace{0.5em}\\
\matX^{+} = \Delta_{j\rightarrow r}\prths{\matX^{-}}, & \matX\in S_r
\end{cases}, ~j,r \in [0,n]
\end{equation}
where $S_r$ is the guard that will trigger the system states' reset transition from the system dynamics $\Sigma_j$ to $\Sigma_r$ via the transition map $\Delta_{j\rightarrow r}$. We show an example of the hybrid system where 4 quadrotors carry a suspended rigid-body payload in Fig.~\ref{fig:multi-hybrid}.

In the following, we first define the guard $S_r$. Let $\matI_p'$ be the index set of the taut cables that become slack. On the other hand, let $\matI_p^{*}$ be the index set of the taut cables whose length tends to increase. Furthermore, let $\matI_z'$ be the index set of the slack cables that reestablish tension. Using these three defined sets, we can immediately obtain that $\matI_p'\subseteq\matI_p,~\matI_p^{*}\subseteq\matI_p$,~ $\matI_z'\subseteq\matI_z$. In addition, the cables in $\matI_p'$ and $\matI_z'$ will change the number of taut cables while those in $\matI_p^{*}$ will not affect them. Therefore, after the system states at $\Sigma_j$ reach the guard region $S_r$ and the corresponding reset finishes, the number of taut cables will be 
\begin{equation}
  r = j + \lvert\matI_z'\rvert - \lvert\matI_p'\rvert,
\end{equation}
where $\lvert*\rvert$ is the size of a set. Now let $d_k$ represent the distance between the $k^{th}$ quadrotor and its attach point on the payload
\begin{equation}
    d_k = \twonorm{\robotpos{k}-\mathbf{p}_{k}}.
\end{equation}
Then the guard $S_r$ can be defined as
\begin{equation}
%     &= \left\{~ , ,,\right\},
    S_r = \left\{\matX~\left\vert
\begin{array}{@{}rl@{}}
&\prths{\matX_{a},\matX_L}\in S_z, a\in\matI_p',\\
&\prths{\matX_{b},\matX_L}\in S_p, b\in\matI_z',\\
&\prths{\matX_{c},\matX_L}\in S_p, c\in\matI_p^{*}
\end{array}\right.
\right\},
\end{equation}
where 
\begin{equation}
\begin{split}
S_z &= \crbrc{\prths{\matX_k,\matX_L}~\lvert~d_k < l_k}, \\
S_p &= \crbrc{ \prths{\matX_k,\matX_L}~\lvert~d_k = l_k,\dot{d}_k>0}.
\end{split}
\label{eq:modelHybridmultiple}
\end{equation}

Further, let's define the transition reset map $\Delta_{j\rightarrow r}$, which has two main parts. The first part is the transition map for the cables in $\matI_p'$. Since when the cable becomes slack, it will just change the dynamical model but not the system state, the transition map for these cables is an identity map. The other is the transition map for the cables in $\matI_z'$ and $\matI_p^{*}$. We model this transition as an inelastic collision. However, it is substantially more complicated than the one in Section~\ref{sec:single-hybrid-model} since the collision will affect not only the velocity of the payload's center of mass but also the angular velocity of the payload.
First we define the impulse acting on the $i^{th}$ robot as $\vecdelta{i}$, where $i\in\matI_z'\bigcup\matI_p^{*}$. Based on the impulse-momentum theorem, the impulse on the robot and payload equals the change in the robot's and payload's momentum
\begin{equation}
    \vecdelta{i} = m_i\prths{\robotvel{i||}^{+} - \robotvel{i||}^{-}},  ~\sum -\vecdelta{i} = \loadmass\prths{\robotvel{L}^{+} - \robotvel{L}^{-}}.\label{eq:multi-load-impulse-translation}
\end{equation}
In addition, the total impulse moment induced by $\vecdelta{i}$ equals the change in the payload's angular momentum
\begin{equation}
   \sum \vecrho{i}\times\prths{-\loadrot^\top\vecdelta{i}}  = \inertiaload(\loadangvel^{+} - \loadangvel^{-}).\label{eq:multi-load-impulse-rotation}
\end{equation}
Since we model the collision as inelastic, the projected velocity of the robot $\robotvel{i||}$ and the corresponding projected attach point velocity $\dot{\mathbf{p}}_{i||}$ along the cable direction should be the same after the collision
\begin{equation}
    \robotvel{i||}^{+} = \dot{\mathbf{p}}_{i||}^{+} =\cablevec{i}\cablevec{i}^\top \dot{\mathbf{p}}_{i}^{+}.\label{eq:multi-equal-attach-vel}  
\end{equation}
By differentiating both sides of \Eqref{eq:attach-rigidbody-payload-geometry} and substituting it into Eqs.~\eqref{eq:multi-load-impulse-translation} and \eqref{eq:multi-load-impulse-rotation} and considering \Eqref{eq:multi-equal-attach-vel}, we can first obtain the linear equations for the linear velocity $\loadvel^{+}$ and angular velocity $\loadangvel^{+}$ of the payload as following
\begin{equation}
    \bar{\matJ}\begin{bmatrix}\loadvel^{+}\\\loadangvel^{+}\end{bmatrix} = \vecb,\label{eq:multi-after-collision}
\end{equation}
where
\begin{equation*}
\begin{split}
    \bar{\matJ} &= 
    \begin{bmatrix}m_L\matI_{3\times3}+\Sigma m_i\cablevec{i}\cablevec{i}^\top & 
    -\Sigma m_i\cablevec{i}\cablevec{i}^{\top}\loadrot\hat{\vecrho{}}_i\vspace{0.5em}\\
    \Sigma m_i\hat{\vecrho{}}_i\loadrot^{\top}\cablevec{i}\cablevec{i}^{\top}&   
    \inertiaload - \Sigma m_i\hat{\vecrho{}}_i\loadrot^{\top}\cablevec{i}\cablevec{i}^{\top}\loadrot\hat{\vecrho{}}_i
    \end{bmatrix}, \\
   \vecb &= \begin{bmatrix}m_L\loadvel^{-}+\Sigma m_i\cablevec{i}\cablevec{i}^\top\robotvel{i}^{-}\vspace{0.5em}\\ 
   \inertiaload\loadangvel^{-} + \Sigma m_i\hat{\vecrho{}}_i\loadrot^{\top}\cablevec{i}\cablevec{i}^{\top}\robotvel{i}
    \end{bmatrix}, \\ 
\end{split}
\end{equation*}
and $\matI_{3\times3}\in\realnum{3\times3}$ is an identity matrix.

\textbf{\textit{Theorem} 1.}The term $\begin{bmatrix}
\loadvel^{+}\\\loadangvel^{+}
\end{bmatrix}$ in the Eq.~(\ref{eq:multi-after-collision}) will always have a non-zero solution for any non-zero $\vecb$.

\textit{Proof.} Let 
\begin{equation}
\begin{split}
   \matA_i =\matA_i^{\top} = \cablevec{i}\cablevec{i}^\top,~
   \matB_i = \hat{\vecrho{}}_i\loadrot^\top,
\end{split}
\end{equation}
Note that $\hat{\vecrho{}}_i$ is a skew-symmetric matrix and we further have
\begin{equation}
   \matB_i^{\top} =  -\loadrot\hat{\vecrho{}}_i.
   %\matA &= m_L\matI_{3\times3}+\Sigma m_i\matA_i, \matB = \Sigma m_i\matA_{i}^{\top}\matB_{i}^{\top}\\
   %\matD & = \inertiaload + \Sigma m_i\matB_i\matA_i\matB_i^{\top}
\end{equation}
Then we can re-write the matrix $\bar{\matJ}$ as 
\begin{equation}
   \bar{\matJ} = 
    \begin{bmatrix} m_L\matI_{3\times3}& \mathbf{0}
    \vspace{0.5em}\\
    \mathbf{0}& \inertiaload  
    \end{bmatrix} + \Sigma m_i\bar{\matJ}_i,\label{eq:reformulated_Jbar}
\end{equation}
where
\begin{equation}
\begin{split}
   \bar{\matJ}_i = \begin{bmatrix} \matA_i & \matA_i\matB_i^\top
    \vspace{0.5em}\\
    \matB_i\matA_i& \matB_i\matA_i\matB_i^{\top}
    \end{bmatrix},
\end{split}
\end{equation}
The first matrix on the right-hand side of \Eqref{eq:reformulated_Jbar} is invertible since both $\loadmass\matI_{3\times3}$ and $\inertiaload$ are positive definite diagonal matrices. Next, we would like to prove that $\bar{\matJ}_i$ is positive semi-definite. If $\bar{\matJ}_i$ is positive semi-definite, then the matrix $\bar{\matJ}$ is positive definite and invertible. And further \textbf{\textit{Theorem} 1} is proven. 
To verify that $\bar{\matJ}_i$ is positive semi-definite, we first know that $\bar{\matJ}_i$ is a symmetric matrix since $\matA_i$ is symmetric. Then for any nonzero vector $\vecx\in\realnum{3}$, we have
\begin{equation}
    \vecx^\top\matA_i\vecx = \prths{\vecx^\top\cablevec{i}}^2 \geq 0 \Rightarrow \matA_i \geq 0. \label{eq:schur_cond_1}
\end{equation}
Hence $\matA_i$ is a positive semi-definite matrix. Let the Singular Value Decomposition (SVD) of $\matA_i$ be
\begin{equation}
    \matA_i = U\Sigma V^{\top},
\end{equation}
where $U$ and $V$ are orthogonal matrices and $\Sigma$ is a diagonal matrix
\begin{equation}
    \Sigma = diag\prths{\sigma_1, \cdots , \sigma_r, 0, \cdots , 0},
\end{equation}
where $\sigma_1 \geq \cdots \geq \sigma_r > 0$ and $r$ is the rank of $\matA_i$. Then the pseudo-inverse of $\matA_i$, denoted as $\matA_i^{\dagger}$, is as in~\cite{gallier2011geometric}
\begin{equation}
    \matA_i^{\dagger} = V\Sigma^{\dagger} U^{\top},
\end{equation}
Then we will have
\begin{equation}
    \matA_i\matA_i^{\dagger}\matA_i = \matA_i,\label{eq:pos-semi-def-property}
\end{equation}
Using \Eqref{eq:pos-semi-def-property}, we can prove the following
\begin{equation}
\prths{\matI_{3\times3} - \matA_i\matA_i^{\dagger}}\matA_i\matB_i^\top =  \matA_i\matB_i^\top - \matA_i\matB_i^\top = 0,~\label{eq:schur_cond_2}
\end{equation}
and
\begin{equation}
\begin{split}
&\matB_i\matA_i\matB_i^{\top} - \prths{\matA_i\matB_i^\top}^{\top}\matA_i^{\dagger}\prths{\matA_i\matB_i^\top}  \\
=&\matB_i\matA_i\matB_i^{\top} - \matB_i\matA_i\matB_i^{\top} = 0 \geq 0.\label{eq:schur_cond_3}
\end{split}
\end{equation}
Then, According to \cite{boyd2004convex}, as we have Eqs.~\eqref{eq:schur_cond_1} - \eqref{eq:schur_cond_3}, we can conclude that $\bar{\matJ}_i\geq 0$. Since $m_i > 0$,  thus the matrix $\bar{\matJ}$ is positive definite, and invertible. $\textbf{Theorem 1}$ is proven $\blacksquare$

From \Eqref{eq:multi-after-collision}, we can obtain
\begin{equation}
    \begin{bmatrix}\loadvel^{+}\\\loadangvel^{+}\end{bmatrix} = \bar{\matJ}^{-1}\vecb.~\label{eq:multi-after-collision-solution}
\end{equation}
We can substitute the solution in the \Eqref{eq:multi-after-collision-solution} into \Eqref{eq:multi-equal-attach-vel} and obtain the solution for $ \robotvel{i||}^{+}$
\begin{equation}
     \robotvel{i||}^{+} = \cablevec{i}\cablevec{i}^\top\prths{\loadvel^{+}-\loadrot\hat{\vecrho{}}_i\loadangvel^{+}}.
\end{equation}
The collision occurs along the cable direction, thus the robot velocity that is vertical to the cable direction $\robotvel{i\perp}^{+}=\robotvel{i\perp}^{-}$. Hence, the robot velocity after the collision $\robotvel{i}^{+} $ is 
\begin{equation}
    \robotvel{i}^{+} = \robotvel{i||}^{+} +\robotvel{i\perp}^{+}
                     = \cablevec{i}\cablevec{i}^\top\prths{\loadvel^{+}-\loadrot\hat{\vecrho{}}_i\loadangvel^{+}} + \robotvel{i\perp}^{-}.~\label{eq:multi-after-collision-solution-final}
\end{equation}
\section{Trajectory Planning and Control}
\subsection{Trajectory Generation}
\label{sec:trajectory_generation}
%\subsubsection{Minimum-kth-derivative Trajectory}
The user can generate a circular trajectory for the payload position in the horizontal $x-y$ plane at the height of $h_c$ like the following: 
\begin{equation}
\loadposdes(t) = \begin{bmatrix}r\cos\frac{2\pi t}{T_c}&r\sin\frac{2\pi t}{T_c}&h_c\end{bmatrix}^\top,
\end{equation}
where $T_c$ is the trajectory's period, and $r$ is the radius of the circular trajectory.
 
In RotorTM, the user can also generate a minimum-kth-derivative~\cite{tang2018aggressive} trajectory made by a series of polynomial trajectories $\loadposdes(t)$ given $m$ waypoints $\robotpos{L,des,0},\cdots,\robotpos{L,des,m}$ that we want the payload to navigate through at time $t_0,\cdots,t_m$ respectively
\begin{equation} 
\loadposdes(t) = 
     \begin{cases}
       \text{$\sum_{n=0}^{N} c_{n1d}t^{n}$} &\quad\text{if $t 	\in \left[t_0,t_1\right],$}\\
        \hspace{10pt}\cdots\\
        \text{$\sum_{n=0}^{N} c_{nid}t^{n}$}&\quad\text{if $t 	\in \left[t_{i-1},t_{i}\right],$}\\
        \hspace{10pt}\cdots\\
        \text{$\sum_{n=0}^{N} c_{nmd}t^{n}$} &\quad\text{if $t 	\in \left[t_{m-1},t_{m}\right],$}
     \end{cases}\label{eqn:poly_spline}
\end{equation}
where $c_{n1d},\cdots,c_{nid},\cdots,c_{nmd}$ are the coefficients of the polynomials. The trajectory is determined by minimizing the total square norm of the $k^{th}$ order derivative %$\displaystyle\frac{d^k\robotpos{L,des}(t)}{dt^k}$ of the $\robotpos{L,des}$
\begin{equation} \argmin_{\loadposdes}
\int_{t_{0}}^{t_{m}} \norm{\frac{d^k\robotpos{L,des}(t)}{dt^k}}^2   \,dt.\label{eqn:cost_integral}
\end{equation}
%\subsubsection{Circular Trajectory}
%with a radius of $r$ and period $T$ at the current height $h$.
%\begin{equation}
%\loadposdes(t) = \begin{bmatrix}r\cos\frac{2\pi t}{T}&r\sin\frac{2\pi t}{T}&h\end{bmatrix}^\top.
%\end{equation}
%The user can also use this trajectory generator for a quadrotor. 
\subsection{Control}\label{sec:controller}
We describe the controllers for the systems in our simulator. %For more details and the proof of the stability of the controllers, the reader can refer to \cite{} 
\subsubsection{Single Quadrotor}
We present the geometric controller for a point-mass cable-suspended payload with a single quadrotor~\cite{sreenath2013cdc}. The desired force acting on the payload is
\begin{align}
\vecF_{des} = &~\prths{m+\loadmass}\prths{\matK_{p}\vece_{\loadpos}+\matK_{d}\vece_{\loadvel}+\matK_{i}\int_0^t\vece_{\loadpos}d\tau}\nonumber\\
&~+\prths{m+\loadmass}\prths{\loadaccdes + \vecg}+ml\prths{\cabledotvec{}\cdot\cabledotvec{}}\cablevec{}, \label{eq:control-single-cable-plpos-control}
\end{align}
where $\matK_{p},\matK_{d},\matK_{i}\in\realnum{3\times3}$ are diagonal gain matrices and $\vece_{\loadpos},\vece_{\loadvel}\in\realnum{3}$ are the payload position and velocity errors defined as
\begin{equation}
\begin{split}
\vece_{\loadpos} = \loadposdes - \loadpos,\\
\vece_{\loadvel} = \loadveldes - \loadvel,~\label{eq:payload-pos-error}
\end{split}
\end{equation}
Since the desired force $\vecF_{des}$ is exerted by the tension cable, it defines the desired tension force vector. Hence the desired tension direction $\cablevec{des}$ can be obtained as
\begin{equation}
 \cablevec{des} = \frac{\vecF_{des}}{\twonorm{\vecF_{des}}}. 
\end{equation}
Then, the thrust $f$ and moment $\matM$ acting on the quadrotor are
\begin{align}
f =& \vecF\cdot\robotrot{}\axis{3}{}\\
\vecF =&\cablevec{}\cablevec{}^\top\vecF_{des}-ml\cablevec{}\times\prths{\matK_{\cablevec{}}\vece_{\cablevec{}}+\matK_{\cablevel{}}\vece_{\cablevel{}}+\prths{\cablevec{}\cdot\cablevel{des}}\cabledotvec{}}\nonumber\\
~&+ml\cablevec{}\times\prths{\cablevec{des}\times\cableddotvec{des}},\label{eq:control-single-cable-qdthrust-control}
\end{align}
\begin{equation}
\begin{split}
\mathbf{M} =&~\matK_{R}\vece_{R} + \matK_{\Omega}\vece_{\Omega} + \robotangvel{}\times\inertia\robotangvel{}\\
&-\inertia\prths{\hat{\robotangvel{}}\robotrot{}^{\top}\robotrot{des}\robotangvel{des}-\robotrot{}^{\top}\robotrot{des}\dot{\robotangvel{}}_{des}},\label{eq:control-single-cable-qdatt-control}
\end{split}
\end{equation}
where $\matK_{R},\matK_{\Omega},\matK_{\cablevec{}},\matK_{\cablevel{}}\in\realnum{3\times3}$ are diagonal gain matrices, $\vece_{R} , \vece_{\Omega}\in\realnum{3}$ are quadrotor's orientation, angular velocity errors defined as
\begin{equation}
\begin{split}
\vece_{\robotrot{}} &= \half\prths{\robotrot{}^{\top}\robotrot{des}-\robotrot{des}^{\top}\robotrot{}}^{\vee}, \\
\vece_{\robotangvel{}} &=  \robotrot{}^{\top}\robotrot{des}\robotangvel{des} - \robotangvel{},\label{eq:robot-angular-error}
\end{split}
\end{equation}
and $ \vece_{\cablevec{}}, \vece_{\cablevel{}}\in\realnum{3}$ are cable direction, cable velocity errors defined as
\begin{equation}
\vece_{\cablevec{}}= \cablevec{des} \times\cablevec{},~
\vece_{\cablevel{}} = \cablevel{} + \cablevec{}\times
\cablevec{}\times\cablevel{des}.\label{eq:cable-error}
\end{equation}
\subsubsection{Multiple Quadrotors with Cables}\label{sec:mult-cable-control}
We introduce the geometric controller for cooperative manipulation of cable-suspended payload with $n$ quadrotors~\cite{Leecdc2014,guanrui2021ral}. The desired forces and moments acting on the payload are
\begin{equation*}
\vecF_{des} = \loadmass\hspace{-2pt}\left(\hspace{-2pt}\matK_{p}\vece_{\loadpos}+\matK_{d}\vece_{\loadvel}+\matK_{i}\hspace{-2pt}\int_0^t\hspace{-5pt}\vece_{\loadpos}d\tau+\loadaccdes + \vecg\hspace{-2pt}\right)
\end{equation*}
\begin{align}
%\label{eq:payloadtotalforceandmoment}
\mathbf{M}_{des} &= \matK_{\loadrot}\vece_{\loadrot}+\matK_{\loadangvel}\vece_{\loadangvel} +\inertiaload\loadrot^{\top}\loadrotdes\loadangaccdes\nonumber\\
&+\prths{\loadrot^{\top}\loadrotdes\loadangveldes}^{\wedge}\inertiaload\loadrot^{\top}\loadrotdes\loadangveldes,\label{eq:control-multi-cable-pl-control} 
\end{align}
where $\matK_{p},\matK_{d},\matK_{i},\matK_{\loadrot}$, $\matK_{\loadangvel}\in\realnum{3\times3}$ are diagonal control gains, $\vece_{\loadpos}, \vece_{\loadvel}$ are the payload position, velocity error defined in \Eqref{eq:payload-pos-error}, $\vece_{\loadrot}, \vece_{\loadangvel}\in\realnum{3}$ are the payload orientation and angular velocity errors defined as
\begin{figure*}[!t]
    \centering
    \includegraphics[width=\textwidth]{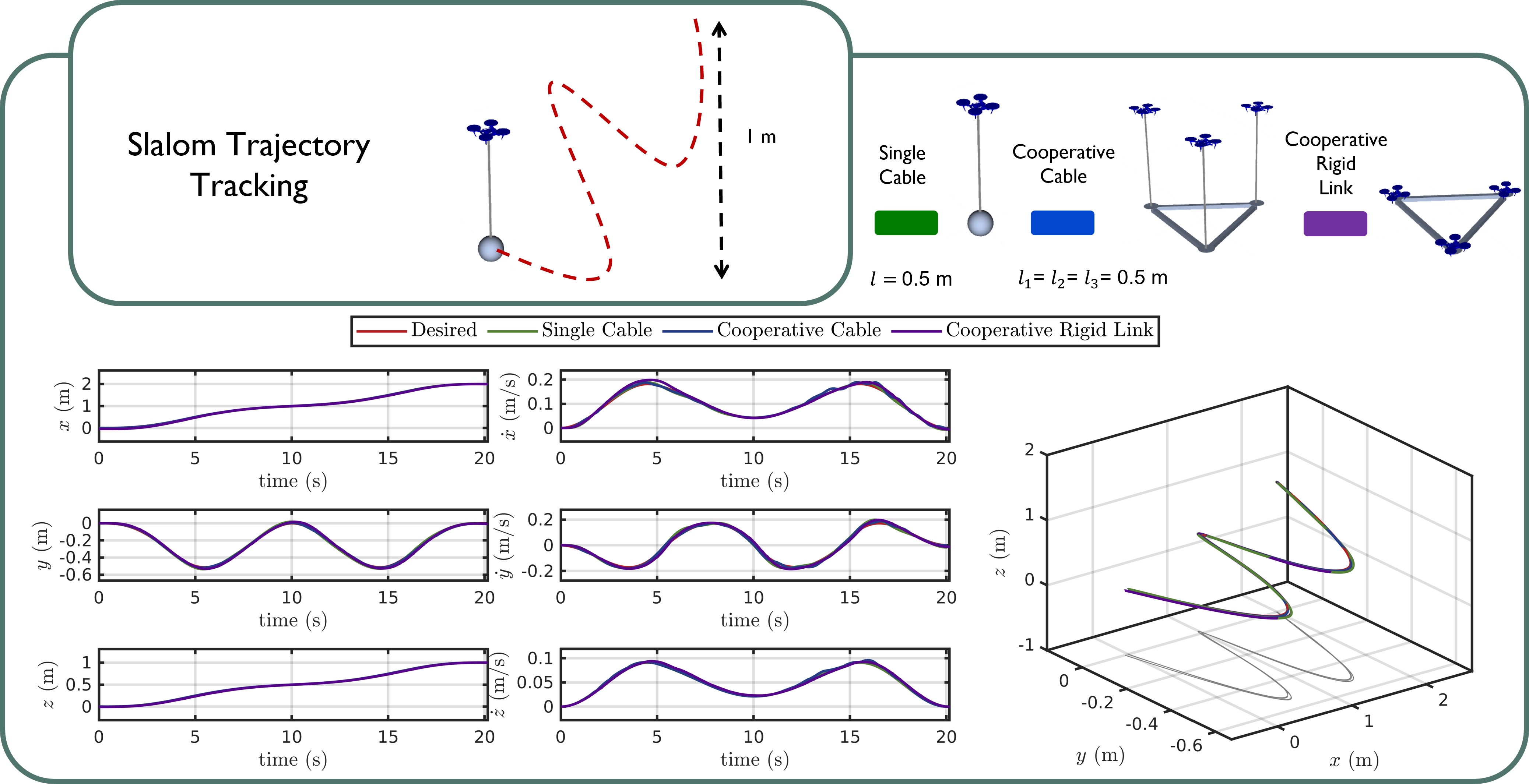}
    \caption{Payload's slalom trajectory tracking results in RotorTM for our sample systems with $t_f = 20s$.}
    \label{fig:min-kth-derivative-traj-tracking}
\end{figure*}
\begin{equation}
\begin{split}
\vece_{\loadrot} &= \half\prths{\loadrot^{\top}\loadrotdes-\loadrotdes^{\top}\loadrot}^{\vee}, \\
\vece_{\loadangvel} &=  \loadrot^{\top}\loadrotdes\loadangveldes - \loadangvel.\label{eq:payload-orientation-errors}
\end{split}
\end{equation}
The force and moment are then distributed to the tensions $\tensiondes{k}$ along each cable as
\begin{equation*}
    \begin{bmatrix}\tensiondes{1}\vspace{-2.5pt}\\
            \vdots\vspace{-2.5pt}\\
            \tensiondes{n}\end{bmatrix} = \text{diag}\left(\loadrot,\cdots,\loadrot\right)\matP^{\top}\left(\matP\matP^{\top}\right)^{-1}\begin{bmatrix}\loadrot^{\top}\vecF_{des}\\
             \mathbf{M}_{des}\end{bmatrix},
\end{equation*}

where $\matP$ is a constant matrix defined as follows: 
\begin{equation}
    \matP = \begin{bmatrix}\matI_{3\times3} & \matI_{3\times3} &\cdots &\matI_{3\times3}\\
            \hatvecrho{1}&\hatvecrho{2}&\cdots&\hatvecrho{n}\end{bmatrix}.
\end{equation}
Based on the desired tension forces, we can obtain the desired direction $\cablevec{k,des}$ and angular velocity $\cableveldes{k}$ of the $k^{th}$ cable link as
\begin{equation}
    \cablevec{k,des} = -\frac{\tensiondes{k}}{\norm{\tensiondes{k}}},~\cableveldes{k} = \cablevec{k,des}\times\cabledotvec{k,des}\,
\end{equation} 
where $\cabledotvec{k,des}$ is the derivative of $\cablevec{k,des}$. The thrust $f_k$ and moments $\matM_k$ acting on the $k^{th}$ quadrotor are
\begin{equation}
\begin{split}
f_k &= \inputforce{k}\cdot\robotrot{k}\axis{3}{}= \left(\inputpara{k}+\inputperp{k}\right)\cdot\robotrot{k}\axis{3}{},\\
\matM_k &= - \matK_{R}\mathbf{e}_{R_k} - \matK_{\Omega}\mathbf{e}_{\Omega_k} + \robotangvel{k}\times\inertia_k\robotangvel{k}, \label{eq:control-multi-cable-qd-control}
\end{split}
\end{equation}
where,  $\matK_{R}$ and $\matK_{\Omega}$ are diagonal control gains, $\mathbf{e}_{R_k}$ and $\mathbf{e}_{\Omega_k}$ are the quadrotor's orientation and angular velocity errors defined in \Eqref{eq:robot-angular-error}, and $\inputperp{k}$ and $\inputpara{k}$ are designed as
\begin{equation*}
    \begin{split}
        \inputpara{k}  =&\,\cablevec{k}\cablevec{k}^{\top}\tensiondes{k} + m_kl_{\mathit{k}}\norm{\cablevel{k}}^2\cablevec{k}  + m_k\cablevec{k}\cablevec{k}^{\top}\veca_{\mathit{k},c},\\
    \inputperp{k}  =&\, m_k l_{\mathit{k}}\cablevec{k}\times\left[-\matK_{\cablevec{k}}\vece_{\cablevec{k}} -\matK_{\cablevel{k}}\vece_{\cablevel{k}} -(\cablevec{k}\cdot\cableveldes{k})\cabledotvec{k} \right]\\
    &-m_k l_{\mathit{k}}\cablevec{k}\times\cablevec{k}\times\cablevec{k}\times\dot{\bm{\omega}}_{k,des}- m_k\cablevec{k}\times\cablevec{k}\times\veca_{\mathit{k},c},
    \end{split}
\end{equation*}
where $\matK_{\cablevec{k}}$, $\matK_{\cablevel{k}}\in\realnum{3\times3}$ are diagonal control gains, $\vece_{\cablevec{k}},\vece_{\cablevel{k}}\in\realnum{3}$ are the cable direction and angular velocity errors of the $k^{th}$ cable similarly defined in \Eqref{eq:cable-error}.
\subsubsection{Multiple Quadrotors with Rigid Links}
We introduce the controller for cooperative manipulation of payload with $n$ quadrotors via rigid links~\cite{LoiannoRAL2018}. The total desired thrust and moment generated by the $n$ quadrotors are
\begin{equation}
\begin{split}
f_c &= m_c\prths{\matK_{p}\vece_{\robotpos{c}}+\matK_{d}\vece_{\robotvel{c}}+\robotacc{c,des} + \vecg}\robotrot{c}\axis{3}{},\\
~\mathbf{M}_{c} &= \matK_{\robotrot{c}}\vece_{\robotrot{c}}+\matK_{\robotangvel{c}}\vece_{\robotangvel{c}} +\inertia_{c}\robotrot{c}^{\top}\robotrot{c,des}\robotangacc{c,des} \\
&+\prths{\robotrot{c}^{\top}\robotrot{c,des}\robotangvel{c,des}}^{\wedge}\inertia_c\robotrot{c}^{\top}\robotrot{c,des}\robotangvel{c,des},\label{eq:control-structure-control}
\end{split}
\end{equation}
where $\matK_{p},\matK_{d},\matK_{i},\matK_{\robotrot{c}}$ and $\matK_{\robotangvel{c}}\in\realnum{3\times3}$ are diagonal control gains and $\vece_{\robotpos{c}}, \vece_{\robotvel{c}}, \vece_{\robotrot{c}}, \vece_{\robotangvel{c}}\in\realnum{3}$ are the structure position, velocity, orientation, and angular velocity errors respectively defined in~\cite{LoiannoRAL2018}. The thrust and moment allocation for each quadrotor is defined by solving an optimization problem as in~\cite{LoiannoRAL2018}. 
%\begin{equation}
%\begin{split}
%    \min_{\matU}\quad&~\matU^\top\matW\matU\\
%    \text{s.t.}\quad& \begin{bmatrix}f_{des}&\matM_{des}^\top      
%    \end{bmatrix}^\top = \begin{bmatrix}
%      \matA_1,\cdots,\matA_n      
%    \end{bmatrix}\matU
%\end{split}
%\end{equation}
%where $\matW$ is a diagonal weight matrix. 

\begin{figure*}[!t]
    \centering
    \includegraphics[width=\textwidth]{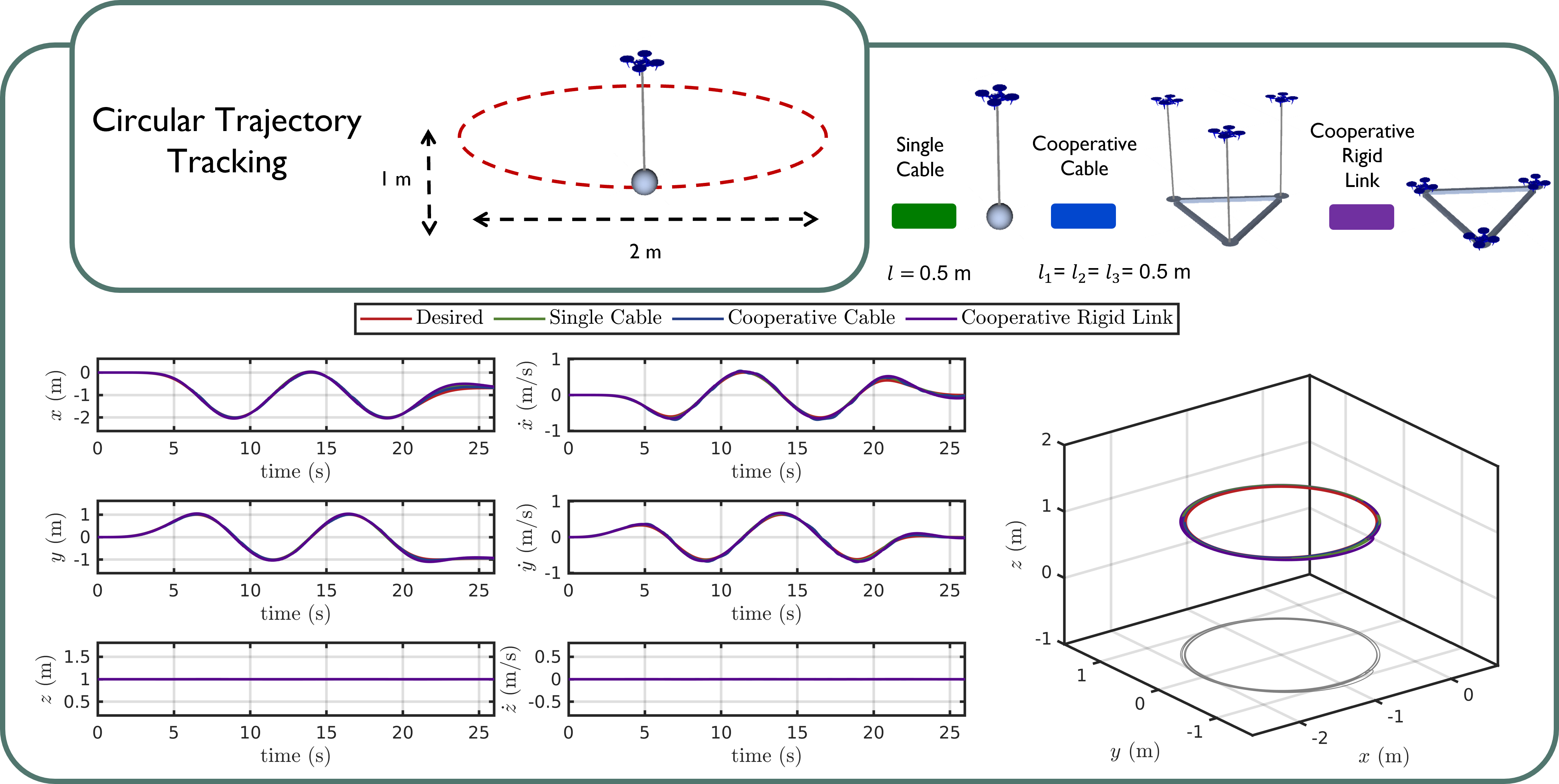}
    \caption{Payload's circular trajectory tracking results in RotorTM for our sample systems with $r=1~\si{m}$, $T_c=10~\si{s}$ and $h_c=1~\si{m}$.}
    \label{fig:circular-traj-tracking}
\end{figure*}
\begin{table*}[!t]
\caption {Simulation Payload Position and Orientation Tracking RMSE in (m)/($^\circ$) with and without feedback loop noise.\label{tab:tracking_error}} 
\centering
%\newcolumntype{s}{}
%\begin{tabular}{\textwidth}{ c c c ccc}
%\begin{tabularx}{\textwidth}{>{\hsize=0.225\hsize}X>{\hsize=0.225\hsize}X>{\hsize=0.19375\hsize}X>{\hsize=0.19375\hsize}X>{\hsize=0.19375\hsize}X>{\hsize=0.19375\hsize}X}
 \begin{tabularx}{\textwidth}{>{\raggedright\arraybackslash}X 
   >{\raggedright\arraybackslash}X 
   >{\centering\arraybackslash}X 
   >{\centering\arraybackslash}X 
   >{\centering\arraybackslash}X 
   >{\centering\arraybackslash}X}
%\begin{tabularx}{\textwidth}{CCCCCC}
    \hline\hline
 \rule{0pt}{2ex} &\rule{0pt}{2ex} &\multicolumn{2}{c}{Slalom}&\multicolumn{2}{c}{Circular}\\
  \cline{3-4}\cline{5-6}
  \rule{0pt}{2ex} & Feedback Noise $\matN$ &$t_m=13 $ s & $t_m =20 $ s & $T = 6$ s & $T = 10$ s \\\hline
  Single Cable    & No Noise             & 0.0407 / NA     & 0.0124 / NA       & 0.115 / NA     & 0.0309 / NA\\
                  & $10^{-3}$            & 0.0397 / NA     & 0.0123 / NA       & 0.110 / NA    & 0.0334 / NA\\\hline
  %                & $10^{-4}$            & 0.0447     & 0.0211       & 0.113    & 0.0331\\\hline
  Multi Cable     & No Noise             & 0.0114 / 0.0388    & 0.00539 / 0.0229     & 0.0439 / 0.113   & 0.0166 / 0.0632\\
                  & $10^{-3}$            & 0.0429 / 0.0743    & 0.0343 / 0.0854      & 0.0711 / 0.164   & 0.0656 / 0.0974\\\hline
  %                & $10^{-4}$            &      &        &     & \\\hline
  Rigid Links     & No Noise             & 0.0426 / 1.541    & 0.0122 / 0.575      & 0.148 / 5.839   & 0.0462 / 2.020\\
                  & $10^{-3}$            & 0.0414 / 1.572    & 0.0126 / 0.568      & 0.155 / 6.205   & 0.0436 / 2.052\\
  %                & $10^{-4}$            & 0.0419     & 0.0155       & 0.144    & 0.0455\\
    \hline\hline
\end{tabularx}
\end{table*}
\section{Results}~\label{sec:experimental_results} 
In this section, we present the results of our simulator. We first show several trajectory tracking results in the simulation with different types of vehicles/payload configurations as well as different types of robots. We obtain the inertial parameters of the payload and the quadrotor platform for the simulation from $\text{SOLIDWORKS}^{\circledR}$ and $\text{AutoDesk Fusion 360}^{\circledR}$ software. Subsequently, we compare the trajectory tracking results obtained in simulation with real-world experiments to fully validate our models as well as control and planning algorithms. 

We conduct real-world experiments at the Agile Robotics and Perception Lab (ARPL), New York University in a $10\times6\times4~\si{m^3}$ flying arena with ``Dragonfly" quadrotor platforms~\cite{LoiannoRAL2017} equipped with a $\text{Qualcomm}^{\circledR}\text{Snapdragon}^{\text{TM}}$ board for onboard computing. They have been used for aerial transportation and manipulation tasks with cable mechanisms and rigid link mechanisms in our previous works~\cite{LoiannoRAL2018,guanrui2021iser,guanrui2021pcmpc,guanrui2023trohumanrobot}. Additionally, in the real-world settings, a Vicon\footnote{\url{www.vicon.com}} motion capture system is leveraged to record the ground truth data for comparison with simulation results at $100~\si{Hz}$.
To perform a fair comparison between our simulator and real-world settings, we use the same inertial parameters of the payload and the quadrotor platform in both scenarios. In addition, we also employ the same controller gains in both cases.  Finally, we follow the same approach to validate our hybrid system and collision models. In this way, we validate our simulator's accuracy, efficacy, and reasonable fidelity with respect to real-world systems. The experiments prove that the simulator can closely mimic real-world systems and can be leveraged to ease the deployments of such complex systems. 

\subsection{Payload Transportation and Manipulation}
\subsubsection{Simulation Results and Analysis}
In the following, we present quantitative and qualitative results from the tests on RotorTM. We test a variety of robot team setups in the simulator including homogeneous quadrotor teams, heterogeneous quadrotor teams, and quadrotor teams with different cable lengths. We also perform several tests with adding the noise in the feedback to the controller. The following results validate the trajectory planning and control in the RotorTM and the RotorTM's capability to simulate different robot setups for aerial transportation and manipulation. 

\textbf{Single robot and homogeneous robot team}:
We show trajectory tracking experiment results in the simulation where a robot or a homogeneous robot team controls the payload to track circular trajectories and slalom trajectories and concurrently to track the desired payload orientation if a rigid-body payload is being carried. In addition, the experiments do not include noise in the feedback loop. We test the following systems: i) a ``Dragonfly" quadrotor carrying a payload via a cable; ii) $3$, $4$, $6$ ``Dragonfly" quadrotors carrying a payload via cables; iii) $3$ ``Dragonfly" quadrotors carrying a payload via rigid links. 

The plots of payload's position and velocity tracking results for the slalom trajectory when $t_f = 20~\si{s}$ and circular trajectory when $r=1~\si{m}$, $T_c=10~\si{s}$ and $h_c=1~\si{m}$ are shown in Figs.~\ref{fig:min-kth-derivative-traj-tracking}-\ref{fig:circular-traj-tracking}. As we can observe, the trajectory generator can generate a smooth trajectory for the payload. The controllers of all three systems can control the payload to track the desired position and velocity with small tracking errors in all Cartesian dimensions. The trajectory tracking tests' position and orientation Root Mean Square Error (RMSE) are reported in Table~\ref{tab:tracking_error}. Our attached multimedia material\footnote{https://youtu.be/jzfEVQ3qlPc} presents the results of $4$ and $6$ quadrotors carrying a payload via cables, where both systems show similar tracking performances.
\begin{figure}[t!]
    \centering
    \includegraphics[width=\columnwidth]{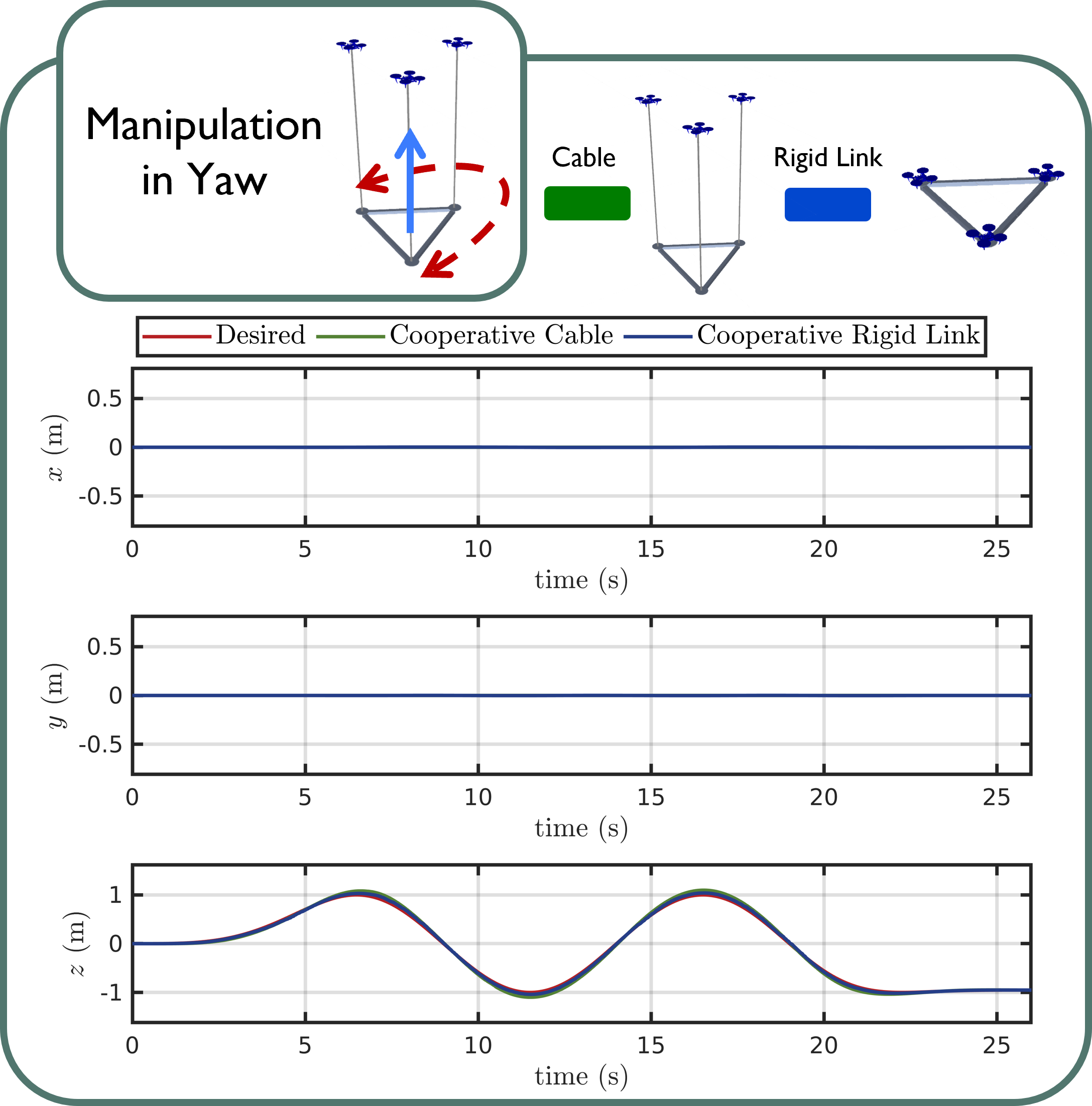}
    \caption{Payload's yaw angle tracking results in RotorTM with 2 different systems: i) 3 "Dragonfly" quadrotors via 1 m cables (green) ii) 3 "Dragonfly" quadrotors via rigid links (blue). }
    \label{fig:yaw-experiments-sim-only}
\end{figure}
We also conduct experiments where a homogeneous team of ``Dragonfly" quadrotors manipulates the payload via both cables and rigid links to rotate around the different axis periodically. As the system with rigid links is a non-holonomic system, the roll and pitch motion are tightly coupled with the motion in the x and y Cartesian directions. Hence in this section, we only show and compare the plots of the payload rotating around the z-axis, i.e, yaw orientation in Fig.~\ref{fig:yaw-experiments-sim-only}, with both rigid links and cable mechanism. In later Section~\ref{sec:TM_realworld_vs_sim}, we will show the results of rotation manipulation in roll, pitch, and yaw using multiple quadrotors via cables in both simulation and real-world experiments. 

In Fig.~\ref{fig:yaw-experiments-sim-only}, the orientation is expressed using the Euler angles ``ZYX" convention, where $\phi$, $\theta$, $\psi$ are the Euler angles correspond to roll, pitch and yaw respectively. As we can observe from the plots, both quadrotor teams with cable and rigid link mechanisms can manipulate the payload to track the desired periodic yaw trajectory closely, validating the manipulation capability of the system.

%We also want to report the simulation calculation run-time scales linearly with respect to the number of robots. The average computational time of one numerical integration cycle using $4^{th}$ order Runge Kutta method for $1$, $3$, $6$ robots are $0.02$ s, $0.05$ s, and $0.075$ s correspondingly.

\begin{figure*}[!t]
    \centering
    \includegraphics[width=\textwidth]{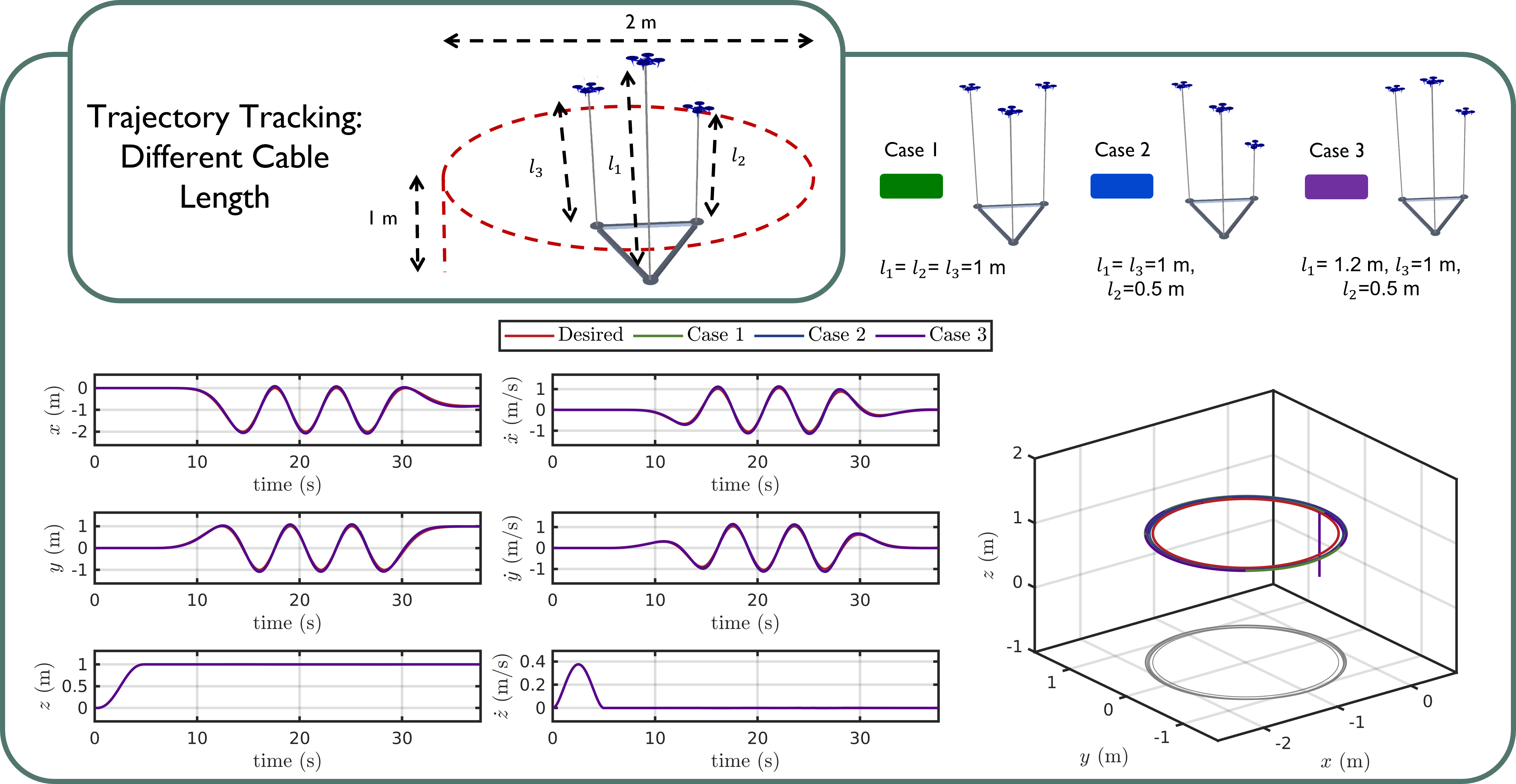}
    \caption{Payload's straight line and circular trajectory tracking results in simulation with homogeneous length and heterogeneous length. The robots first takeoff the payload to $1~\si{m}$ of height and transport the payload to track the circular trajectory with $r=1~\si{m}$, $T_c=6~\si{s}$ and $h_c=1~\si{m}$.}
    \label{fig:different-length-cables}
\end{figure*}

\textbf{Different cable length}:
We further conduct payload trajectory tracking experiments in simulation using different cable lengths. In this set of experiments, we use a homogeneous robot team that consists of ``Dragonfly" quadrotors only. We tested 3 cases i) all 3 quadrotors use $1~\si{m}$ cable ; ii) 2 quadrotors use $1~\si{m}$ cable while 1 quadrotor use $0.5~\si{m}$ cable; iii) 1 quadrotor use $1.2~\si{m}$ cable, 1 quadrotor use $1~\si{m}$ cable and 1 quadrotor use $0.8~\si{m}$ cable. 

The plots of the payload's position and velocity tracking results for a circular trajectory when $r=1~\si{m}$, $T_c=6~\si{s}$ and $h_c=1~\si{m}$ are shown in Fig.~\ref{fig:different-length-cables}. The controllers of all three systems can control the payload to track the desired position and velocity with minor tracking errors in all Cartesian dimensions. This demonstrates that our simulator can deploy and simulate a robot team with different cable lengths for aerial transportation and manipulation.

\begin{table}[t!]
  \centering
  \caption{Robot Types and Parameters.}\label{tab:robot_parameters}
%  \begin{tabularx}{\textwidth}{>{\hsize=0.225\hsize}X|>{\hsize=0.225\hsize}c|>{\hsize=0.225\hsize}c|>{\hsize=0.225\hsize}X}
\begin{tabularx}{\columnwidth} { 
   >{\raggedright\arraybackslash}X 
   >{\centering\arraybackslash}X 
   >{\centering\arraybackslash}X 
   >{\centering\arraybackslash}X}
%  \begin{tabular}{ | c m{5em} | m{5cm}| m{5cm} | m{5cm}| } 
    \hline\hline
    & Dragonfly~\cite{LoiannoRAL2018} & Hummingbird~\cite{RSS2013Koushil} & Race \\ 
    &\begin{minipage}{.25\columnwidth}
    %\centering
      \includegraphics[width=20mm, height=15mm]
      {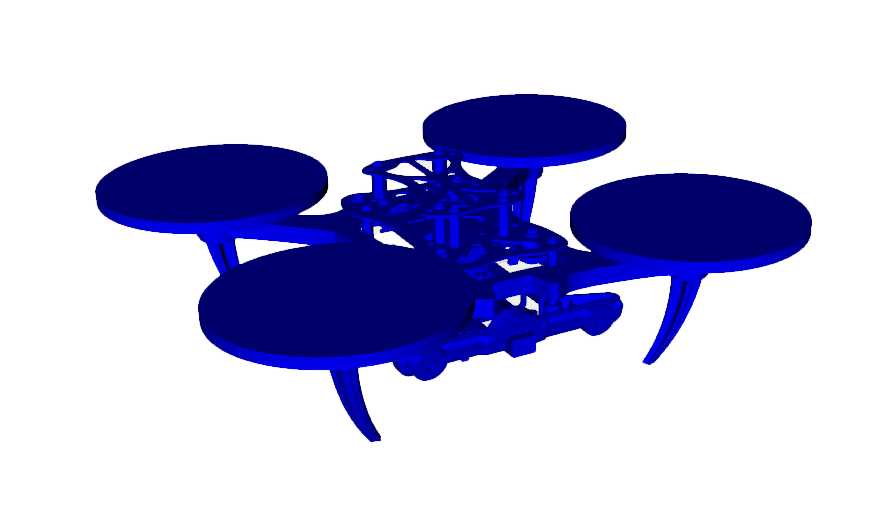}
    \end{minipage}
    &\begin{minipage}{.25\columnwidth}
    %\centering
      \includegraphics[width=20mm, height=15mm]
      {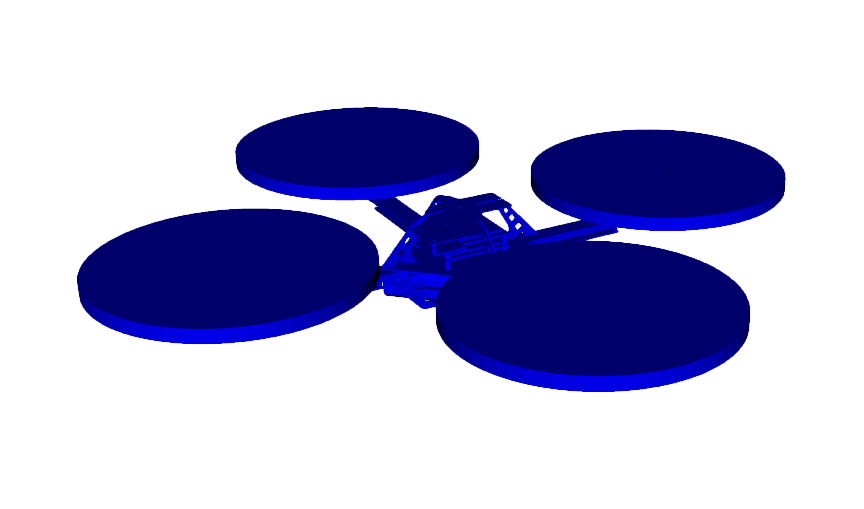}
    \end{minipage}
    & \begin{minipage}{.25\columnwidth}
    %\centering
      \includegraphics[width=20mm, height=15mm]
      {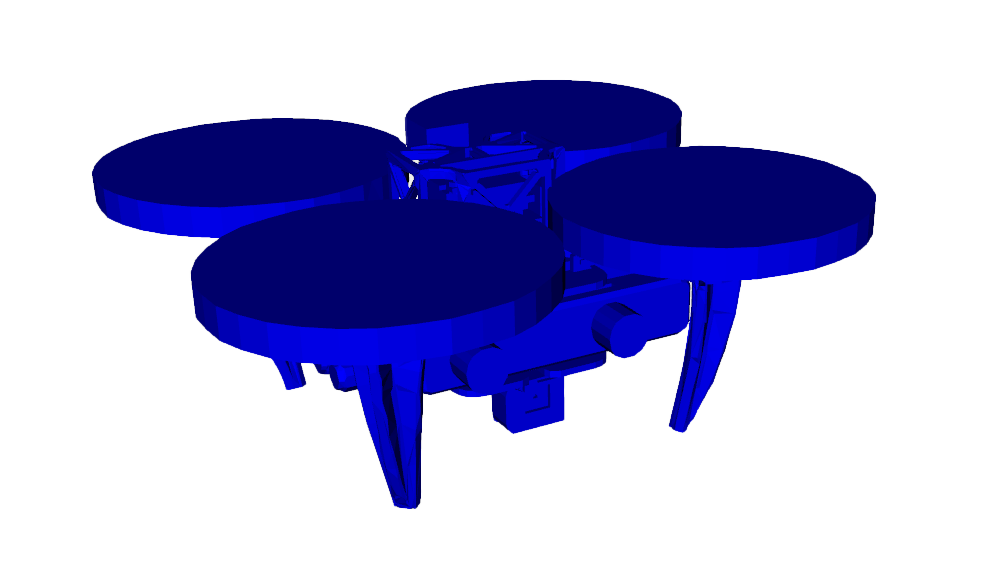}
    \end{minipage}
    \\\hline 
    mass (\si{kg}) & 0.25 & 0.5 & 0.95 \\
arm length (\si{m})& 0.1075 & 0.17 & 0.10125 \\
$J_{xx}$ (\si{kg\cdot m^2})& 0.601$\times10^{-3}$ & 2.64$\times10^{-3}$ & 3.0$\times10^{-3}$ \\
$J_{yy}$ (\si{kg\cdot m^2})& 0.589$\times10^{-3}$ & 2.64$\times10^{-3}$ & 3.0$\times10^{-3}$ \\
$J_{zz}$ (\si{kg\cdot m^2})& 1.076$\times10^{-3}$ & 4.96$\times10^{-3}$ & 4.0$\times10^{-3}$ \\
max motor speed (RPM) & 16400 & 7500 & 23000 \\
min motor speed (RPM) & 5500  & 1500 & 5500 \\
%motor_coefficents: 5.55e-8
%motor_coefficients: 1.21702136e-8
%motor_coefficients: 3.9408732e-9 # N/rpm^2
    \hline\hline
  \end{tabularx}
  %\end{tabular}
\end{table}

\textbf{Heterogeneous robot team}: In the following, we conduct payload trajectory tracking experiments in simulation using cooperative heterogeneous robot teams via cable or rigid link mechanisms. In these experiments, the heterogeneous robot team will transport the payload to take off to $1$ m height and track the circular trajectories, simultaneously controlling the payload's orientation to the desired orientation. The heterogeneous robot team tested in these experiments consists of 1 ``Dragonfly"~\cite{LoiannoRAL2017,guanrui2021pcmpc,guanrui2021iser}, 1 ``Hummingbird"~\cite{tang2018aggressive} and 1 ``Race" quadrotor (customized quadrotor platform in ARPL with 5-inch propeller), with parameters shown in Table~\ref {tab:robot_parameters}.
\begin{figure*}[!t]
    \centering
    \includegraphics[width=\textwidth]{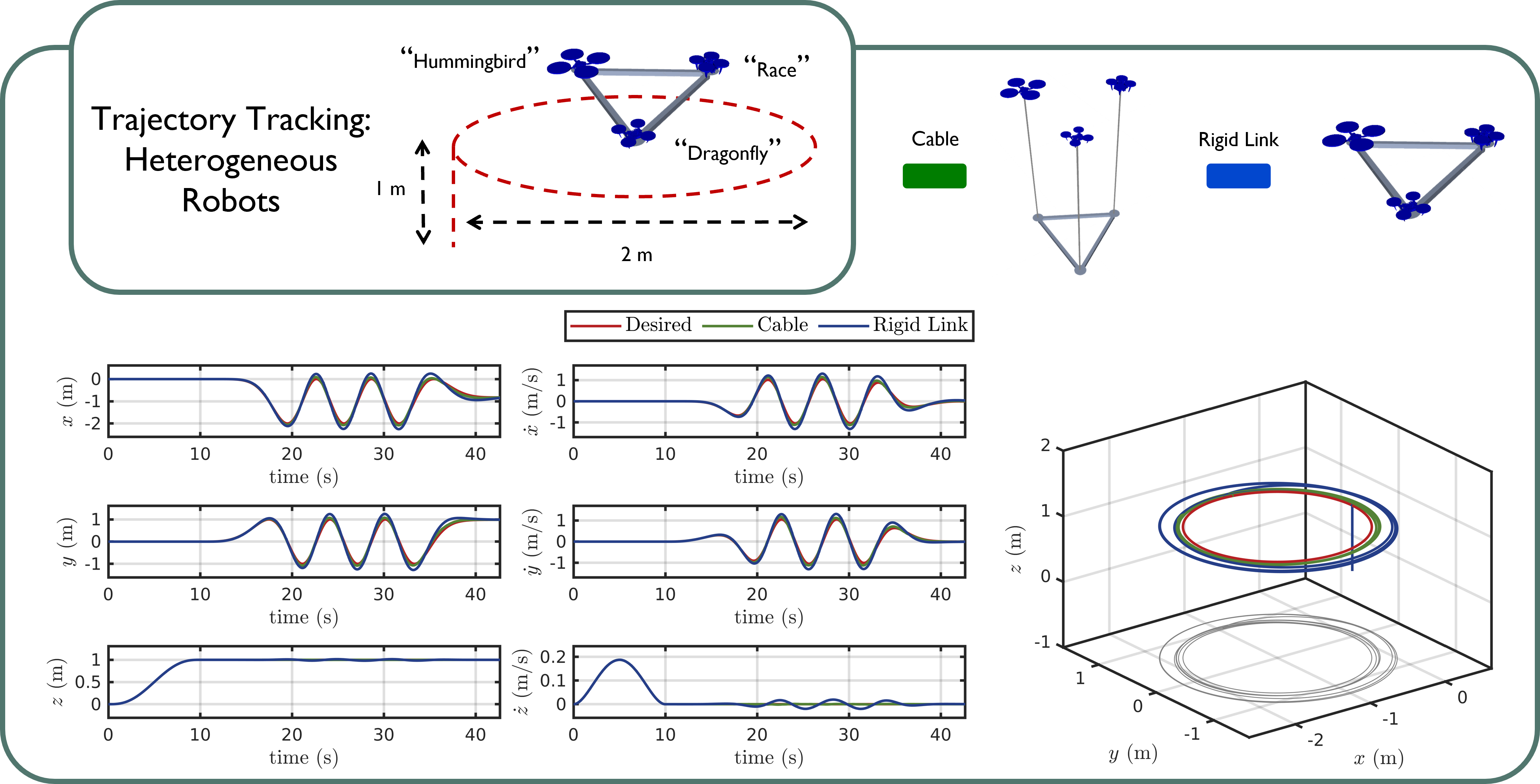}
    \caption{Payload's straight line and circular trajectory tracking results in simulation with heterogeneous robots. The robots first takeoff the payload to $1~\si{m}$ height and transport the payload to track the circular trajectory with $r=1~\si{m}$, $T_c=6~\si{s}$ and $h_c=1~\si{m}$.}
    \label{fig:heterogeneous-robots}
\end{figure*}

The plots of payload's position and velocity tracking results for circular trajectory when $r=1~\si{m}$, $T_c=6~\si{s}$ and $h_c=1~\si{m}$ are shown in Fig.~\ref{fig:heterogeneous-robots}. The controllers of all three systems can maneuver the payload to track the desired position and velocity with minor tracking errors in all Cartesian dimensions. This show that our simulator can deploy and simulate heterogeneous robots with different dynamic properties for aerial transportation and manipulation.

\begin{figure*}[!t]
    \centering
    \includegraphics[width=\textwidth]{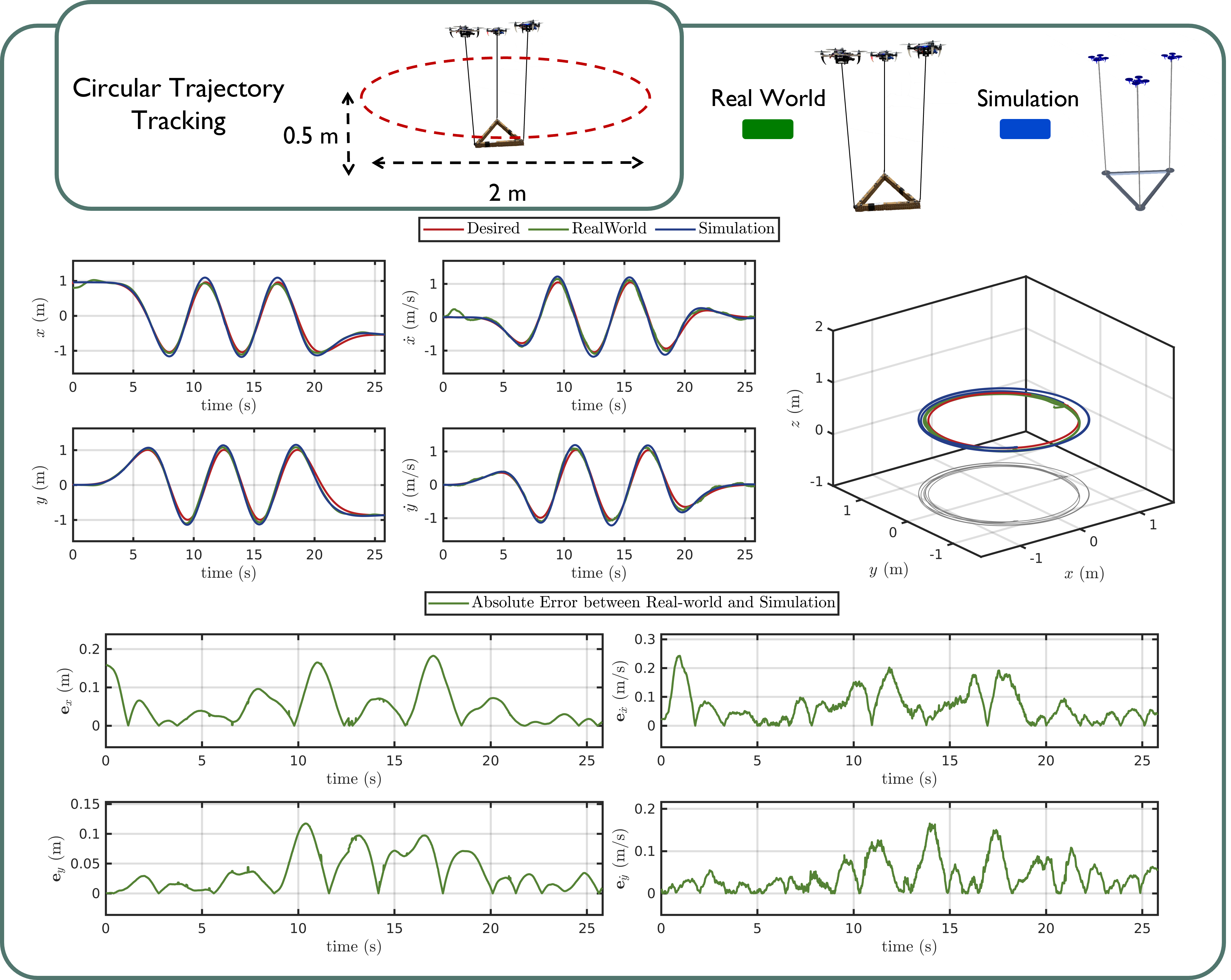}
    \caption{Comparison between real-world and simulation experiments. In this case, we show the tracking results in both real world and simulation of three ``Dragonfly" quadrotors carrying a suspended triangular payload via $1$ m cables to track a circular trajectory with a period of 6s and a radius of 1m.}
    \label{fig:circular_traj_tracking_cable}
\end{figure*}
\textbf{Noise in the feedback loop}: Finally, we compare payload trajectory tracking results in simulation with and without the presence of noise in the feedback loop. In this set of experiments, we use a homogeneous robot team that consists of ``Dragonfly" quadrotors only. We test the following systems: i) a quadrotor transporting a payload via a cable; ii) $3$ quadrotors carrying a payload via cables; iii) $3$ quadrotors carrying a payload via rigid links. 

As shown in Fig.~\ref{fig:control_block_diagram}, we can add noise on the system state vector $\matX(t)$ from the state integration block. Here, we add zero-mean Gaussian noise $\matN$ with zero mean and covariance $Q_{\matN}$ as
\begin{equation}
    \bar{\matX}(t) = \matX(t) \oplus \matN, \,~\matN \sim\mathcal{N}\prths{\mathbf{0},Q_{\matN}},
\end{equation}
where $\oplus$ performs the mapping between the manifold and the tangent space such that we can use them similarly to $+$ in the vector space. For most of the states in $\matX(t)$, such as position, velocity, and angular velocity, we sample the noise from the Gaussian distribution in the vector space and directly add it component-wise. For the orientation $\vecq$ in $\matX(t)$, we sample the noise $\boldsymbol{\sigma}_q$ from the Gaussian distribution in the vector space and map it to the quaternion manifold via the corresponding exponential function~\cite{sola2017QuaternionKF} and add it to the orientation $\vecq$ as
\begin{equation}
  \vecq \oplus \boldsymbol{\sigma}_q := \vecq \otimes \exp_{\vecq} \prths{\frac{\boldsymbol{\sigma}_q}{2}},
\end{equation}
where $\otimes$ denotes the quaternion product.
We set the standard deviation of the Gaussian distribution as $10^{-3}$ and test the same trajectories as those we report in Table~\ref{tab:tracking_error}. The trajectory tracking tests' position and orientation Root Mean Square Error (RMSE) are reported in Table~\ref{tab:tracking_error} accordingly. The controllers of all three systems can control the payload to track the desired Cartesian trajectory and orientation with minor tracking errors and showing similar performance and accuracy as in the case of absence of noise.

\subsubsection{Real-world vs. Simulation Comparison}
\label{sec:TM_realworld_vs_sim}
In this section, we compare simulation and real-world experiments to validate our controller, trajectory planner, and simulator models. We use the same ``Dragonfly" quadrotor platform in both simulation and real-world environments, which has been used in our previous works~\cite{guanrui2021iser,LoiannoRAL2018,guanrui2021ral,guanrui2023trohumanrobot}. The state feedback and ground truth in the real-world experiments are provided by the Vicon system at $100~\si{Hz}$.

To ensure a fair comparison between simulation and real-world settings, we use the same control gains for both payload controller and robot controllers in both simulation and real-world scenarios. Additionally, we also include the plots of absolute error between the real-world data and simulation data in each experiment, which show the fidelity of the simulator. 
\begin{figure*}[!t]
    \centering
    \includegraphics[width=\textwidth]{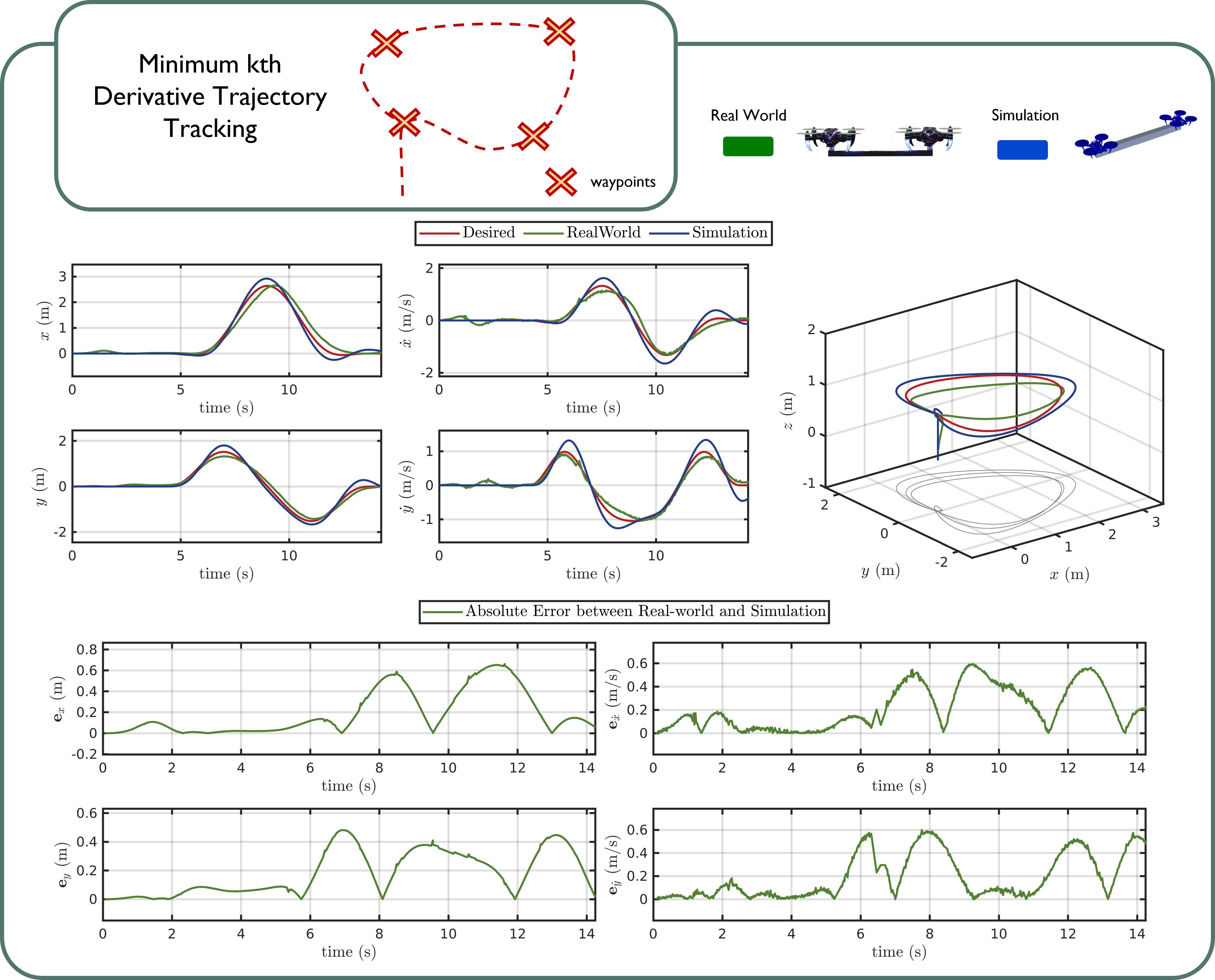}
    \caption{Comparison of real-world and simulation experiments
of two quadrotors carrying a bar payload through rigid links
to track a trajectory generated by the minimum k-th derivative trajectory generators.}
\label{fig:circular_traj_tracking_rigid_links}
\end{figure*}
\begin{table}[!t]
%\vspace{-10pt}
\caption {Simulation vs Real-World Payload Transportation Task Position Tracking  RMSE (m).\label{tab:tracking_error_vs_real}} 
\centering
%\newcolumntype{s}{}
\begin{tabularx}{\columnwidth}{>{\hsize=0.225\hsize}X>{\hsize=0.1\hsize}X>{\hsize=0.375\hsize}X>{\hsize=0.375\hsize}X}\hline\hline
 \rule{0pt}{2ex} & &Simulation & Real-World \\\hline
 Multi Cable& $x$    & 0.0699    &  0.0703 \\
            & $y$    & 0.0716    &  0.0862 \\
            & $z$    & 0.0140    &  0.0349 \\\hline
 Multi Rigid& $x$    & 0.115     &  0.170 \\
            & $y$    & 0.116     &  0.111 \\
            & $z$    & 0.0875      & 0.0929 \\    
    \hline\hline
\end{tabularx}
\vspace{-10pt}
\end{table}

\textbf{Transportation}. In the transportation experiments, we test and compare the results of i) 3 ``Dragonfly" quadrotors transporting a suspended triangular payload while tracking a circular trajectory generated by the circular trajectory generator, ii) 2 ``Dragonfly" quadrotors transporting a rigidly attached bar payload to track a trajectory generated by minimum-$k^{th}$-derivative trajectory. In case i), we conduct the real-world flight in our flying arena and collect the corresponding data, whereas in case ii), we leverage real-world experiment data of our previous work~\cite{LoiannoRAL2018}. The experiment results are shown in Figs.~\ref{fig:circular_traj_tracking_cable}-\ref{fig:circular_traj_tracking_rigid_links} and the RMSE of trajectory position tracking are reported in Table~\ref{tab:tracking_error_vs_real}. 

In Table~\ref{tab:tracking_error_vs_real}, we notice that the tracking RMSE in the real world is about $0.01~\si{m}~(1~\si{cm})$ larger than the simulation ones. This is reasonable because there are some additional unmodeled effects like communication delays or disturbances in the real-world environment which are not simulated in RotorTM. Based on the results, we can conclude that the real-world and simulation experiments have similar tracking errors in both payload's position and velocity, therefore validating our simulator. 

\begin{figure*}
    \centering
    \includegraphics[width=\textwidth]{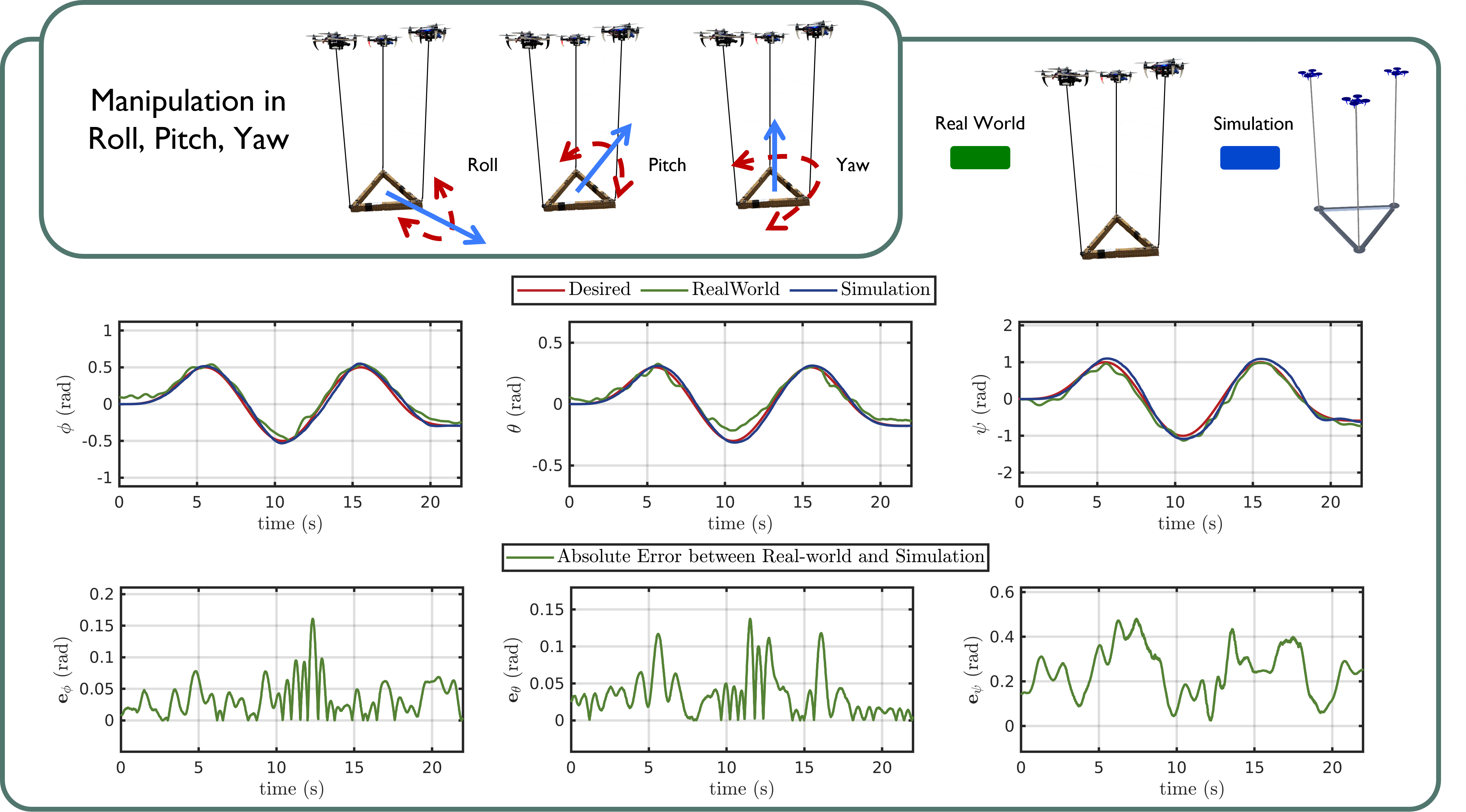}
    \caption{Comparison between real-world and simulation experiments. In this case, we show the tracking results in both real world and simulation of three ``Dragonfly" quadrotors carrying a suspended triangular payload via $1$ m cables to cooperatively rotate the payload’s yaw, pitch, and roll orientation periodically.}
    \label{fig:yaw_experiment_realworld_vs_sim}
\end{figure*}
\begin{table}[!t]
%\vspace{-10pt}
\caption {Simulation vs Real-World Payload Manipulation Task Tracking RMSE (degrees) using multiple quadrotors with cable mechanisms \label{tab:manipulation_tracking_error_vs_real}} 
\centering
%\newcolumntype{s}{}
\begin{tabularx}{\columnwidth}{>{\hsize=0.33\hsize}X>{\hsize=0.33\hsize}X>{\hsize=0.33\hsize}X}\hline\hline
 \rule{0pt}{2ex} Traj. Type   & Simulation & Real-World \\\hline
 $\phi$   & 1.992    &  3.648 \\
 $\theta$  & 1.206    &  2.635 \\
 $\psi$    & 5.837    &  6.888 \\
    \hline\hline
\end{tabularx}
\vspace{-10pt}
\end{table}
\begin{figure}[t!]
    \centering
    \includegraphics[width=\columnwidth]{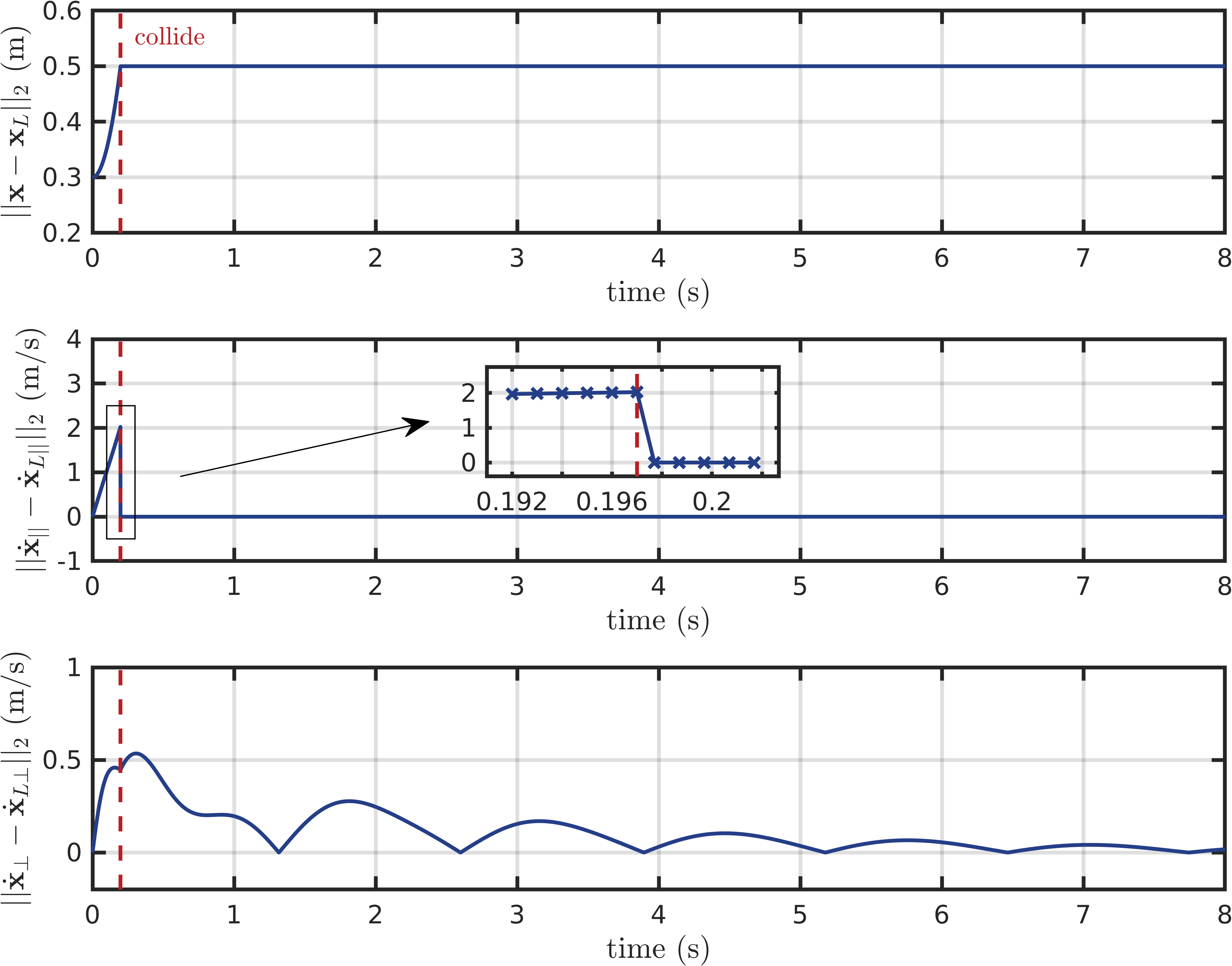}
    \caption{Simulation results of a single robot with cable-suspended payload. The cable transition from slack to taut condition at the moment denoted by the red dashed line, and the collision between the robot and the payload occurs.}
    \label{fig:collide}
    \vspace{-10pt}
\end{figure}
\textbf{Manipulation}.
We also conduct the payload manipulation task in both simulation and real-world environments for comparison. We test that the ``Dragonfly" robot team cooperatively rotates the payload’s yaw, pitch, and roll orientation periodically via cables. The orientation tracking RMSE along the periodic rotation direction in degrees is reported in Table~\ref{tab:manipulation_tracking_error_vs_real}. The periodic orientation trajectory tracking results in all 3 dimensions have been shown in Fig.~\ref{fig:yaw_experiment_realworld_vs_sim}. 

Table~\ref{tab:manipulation_tracking_error_vs_real} shows that the orientation tracking RMSE in the real world generally is about $1-2$ degrees larger than the simulation ones. These are reasonably small errors. Based on the results, we can conclude that the real-world and simulation experiments have similar tracking errors in the payload's orientation, validating our simulator. 

\begin{figure}[t!]
    \centering
    \includegraphics[width=\columnwidth]{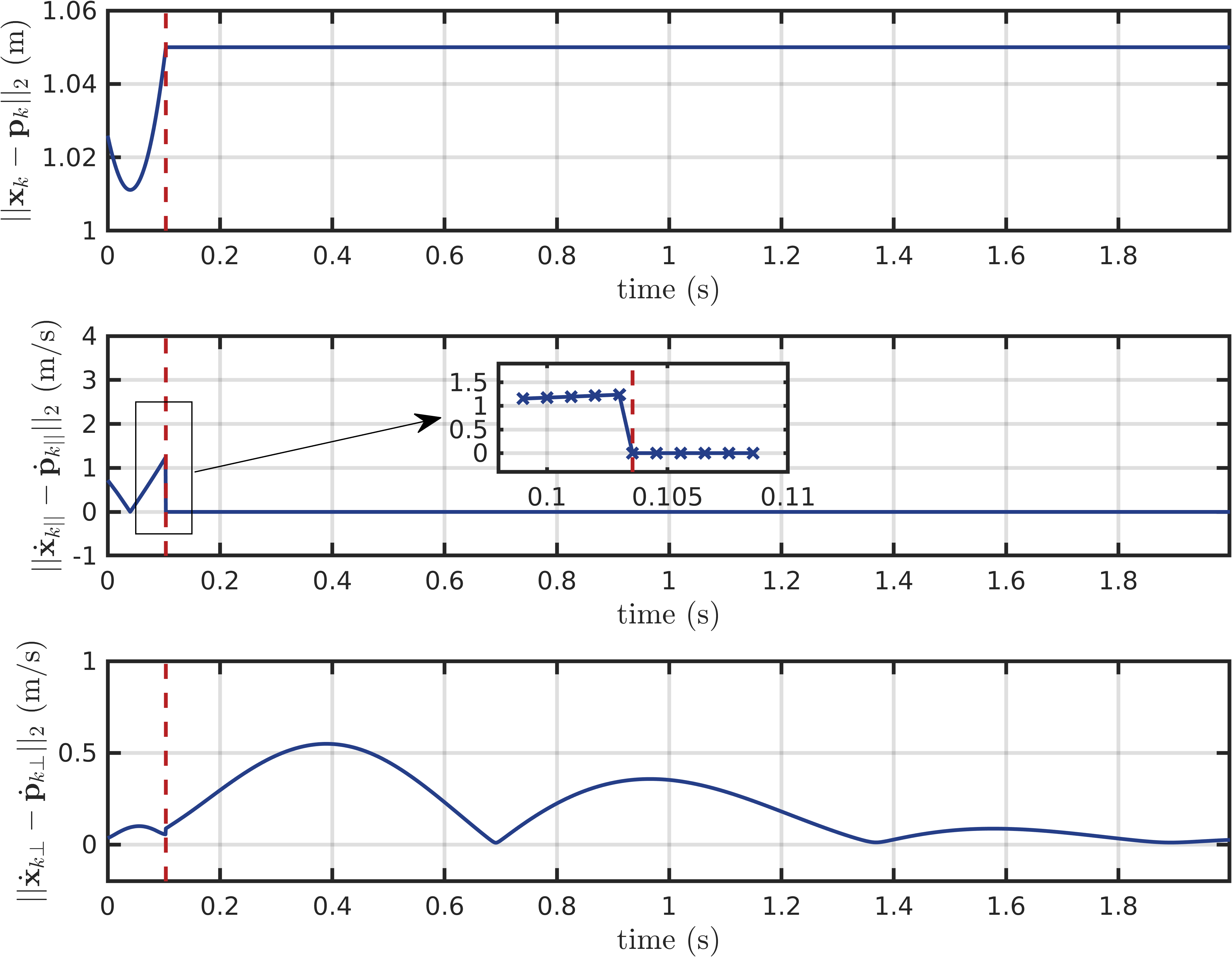}
    \caption{Simulation results of multiple robots with cable suspended payload. The cable at the front robot transitions from slack to taut condition at the moment denoted by the red dashed line, and the collision between the robot and the payload occurs.}
    \label{fig:multi_robot_collide}
    \vspace{-10pt}
\end{figure}
\begin{figure*}[t]
    \centering
        \subfigure[Before poking the payload, the cable in front is taut.\label{fig:real-world-collide-demo-before-poking}]{
    \includegraphics[width=0.32\textwidth]{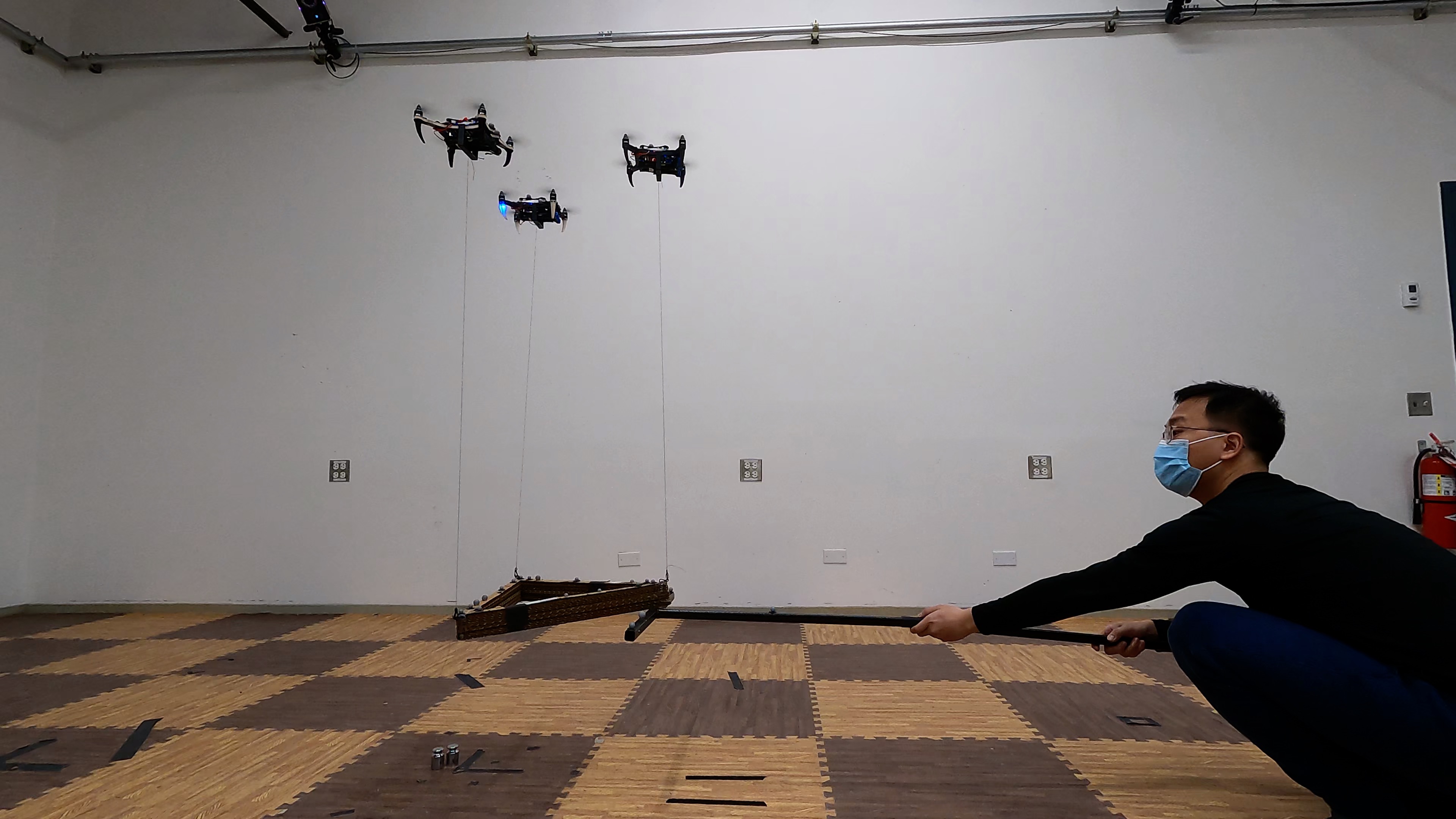}}
        \subfigure[During poking the payload, the cable in front becomes slack.\label{fig:real-world-collide-demo-during-poking}]{
    \includegraphics[width=0.32\textwidth]{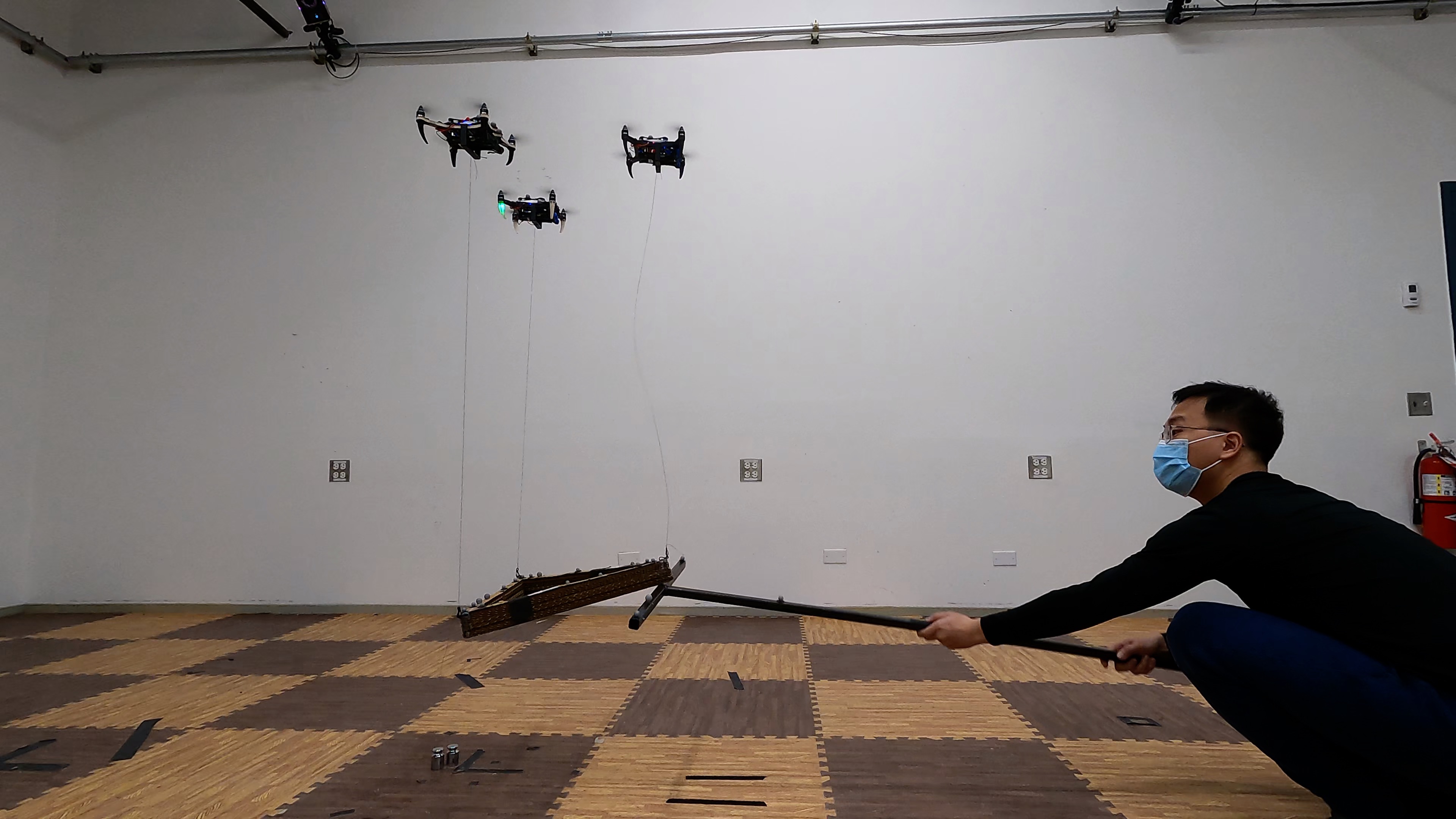}}
        \subfigure[After poking the payload, the cable in front returns taut.\label{fig:real-world-collide-demo-after-poking}]{
    \includegraphics[width=0.32\textwidth]{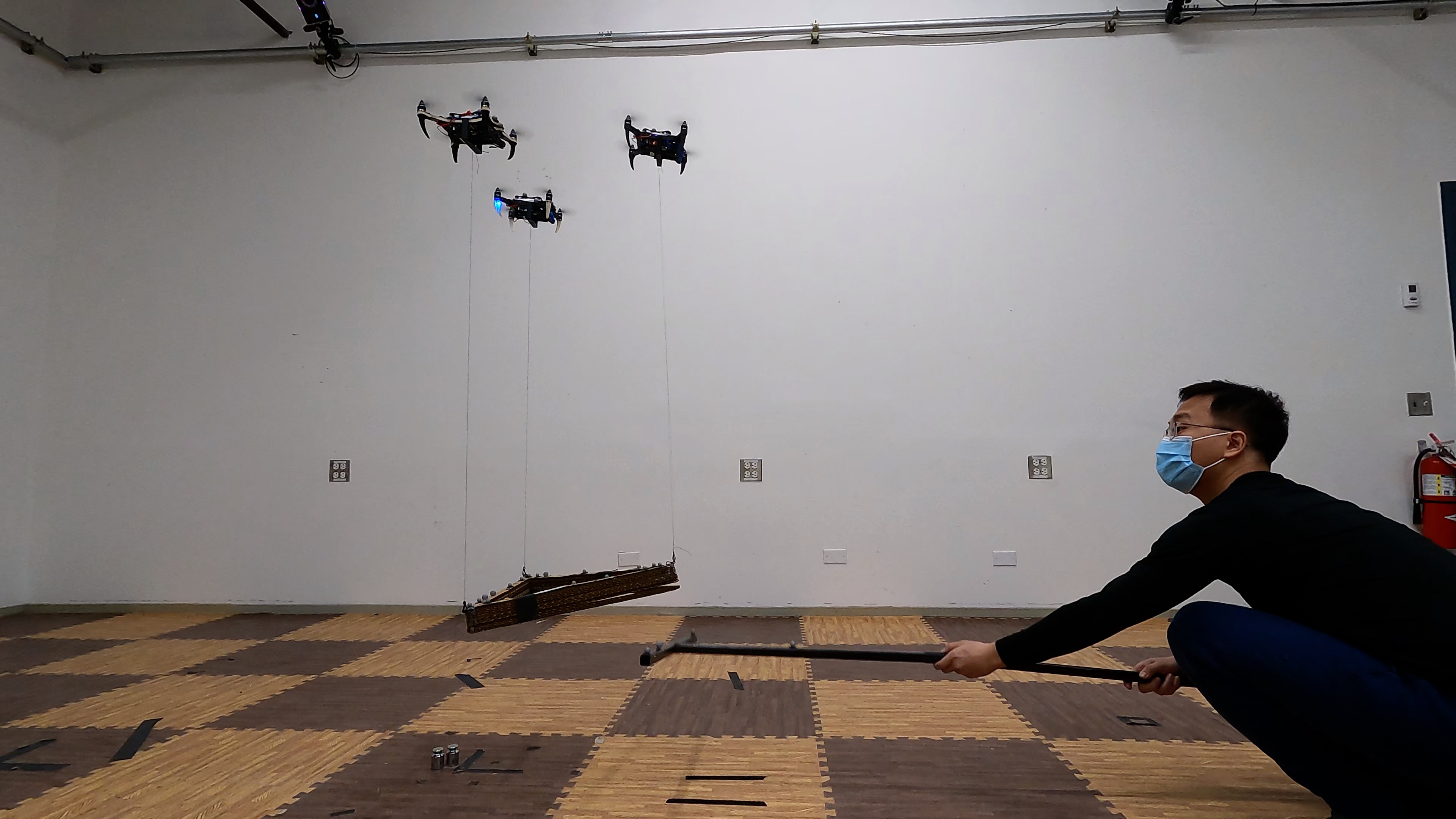}}
    \caption{Real-world experiments where we poke the payload with a wand at one of the payload's corners. The front cable becomes slack and reestablishes tension later. We record the system states using the Vicon motion capture system and initialize our simulator with the recorded state when the cable is slack. A video of both real-world experiment and simulation can be found at this link: \url{https://youtu.be/jzfEVQ3qlPc}} \label{fig:real-world-collide-demo}
    \vspace{-10pt}
\end{figure*}
\begin{figure*}
\centering
    \includegraphics[width=0.8\textwidth]{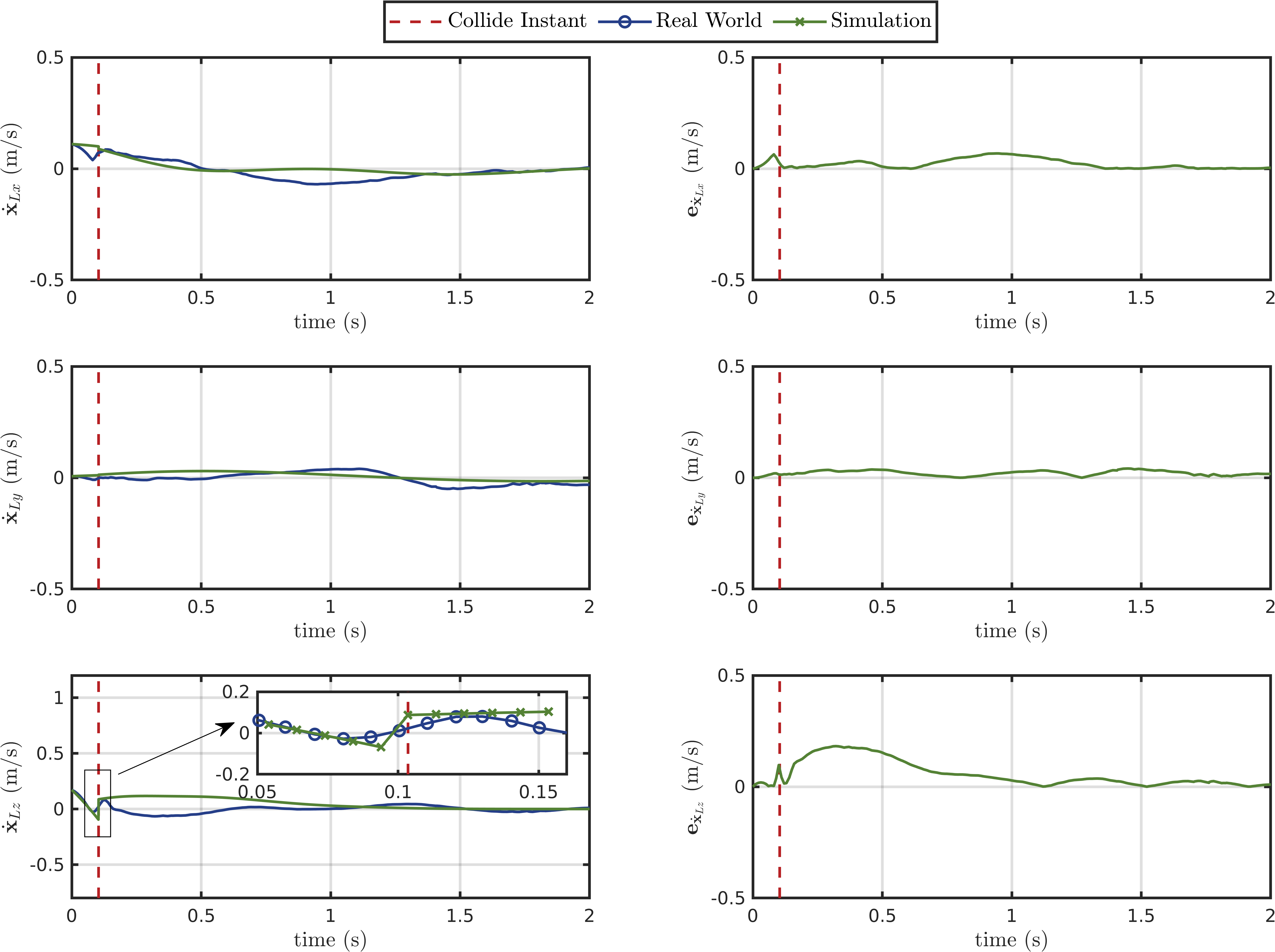}
    \caption{Payload velocity (real-world vs simulation) and the corresponding error plots. As shown in the figure, we observe the payload linear velocity in simulation reaches similar values compared to the ones in the real world, proving the simulator can simulate the hybrid system transition with fidelity.\label{fig:multi-collide-payload-vel}}
\end{figure*}
\begin{figure*}
\centering
    \includegraphics[width=0.8\textwidth]{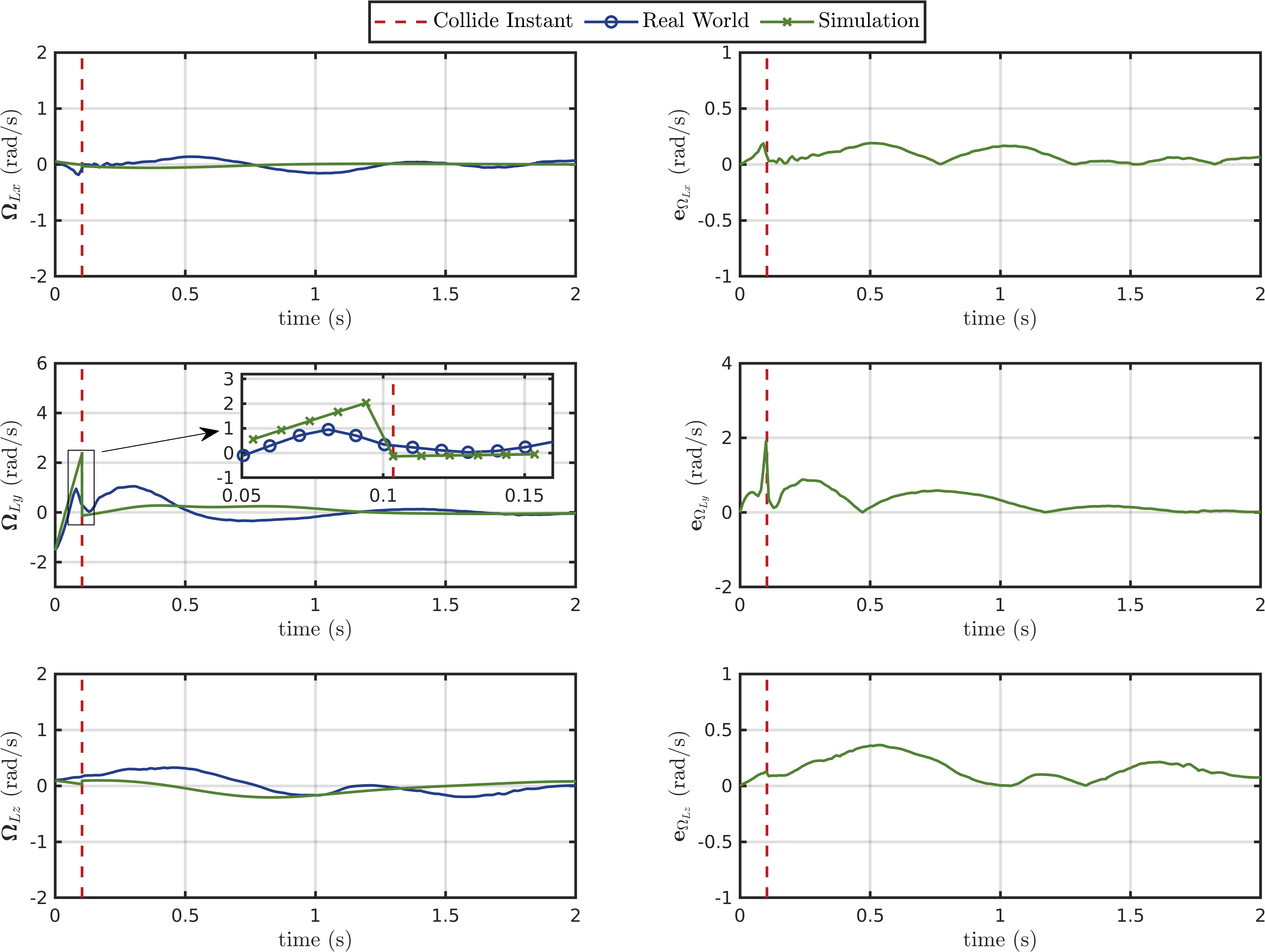}
    \caption{Payload angular velocity (real-world vs simulation) and the corresponding error plots. As shown in the figure, we observe the payload angular velocity in simulation reaches similar values compared to the ones in the real world, proving the simulator can simulate the hybrid system transition with fidelity. \label{fig:multi-collide-payload-ang-vel}}
\end{figure*}

\subsection{Hybrid System}
\subsubsection{Simulation Results}
We show and validate the inelastic collision simulation results of the system with a single ``Dragonfly" quadrotor transporting a point-mass payload via a $0.5~\si{m}$ cable in Fig.~\ref{fig:collide}. We release the quadrotor at the height of $1~\si{m}$ and the payload with a distance of $0.3~\si{m}$ from the quadrotor, which is shorter than the taut cable length $0.5~\si{m}$. Furthermore, the direction from the payload to the quadrotor is $30^{\circ}$ from the $\axis{3}{}$ of the world frame $\worldf$.
As we can observe in Fig.~\ref{fig:collide}, the collision happens when the distance between the payload and quadrotor $\norm{\robotpos{}-\loadpos}_2$ reaches $0.5~\si{m}$ denoted by the red dotted line. After the collision, the norm of the relative velocity between the quadrotor and payload projected on the cable $\norm{\robotvel{||}-\robotvel{L||}}_2$ becomes $0$. And the relative velocity orthogonal to the cable direction remains the same when the collision occurs. Subsequently, the controller will recover the payload position to its desired hovering position. 

Similarly, we show and validate the inelastic collision simulation results of the system with three ``Dragonfly" quadrotors carrying a rigid-body payload via $1.05~\si{m}$ cables as shown in Fig.~\ref{fig:multi_robot_collide}. We initialize the system in the simulator with recorded states from Vicon in the corresponding real-world experiment when the front cable becomes slack. This ensures a fair comparison with real-world data to fully validate our model. 
As shown in Fig.~\ref{fig:multi_robot_collide}, the collision happens when the distance between the payload attach point and quadrotor $\norm{\robotpos{k}-\mathbf{p}_{k}}_2$ reaches $1.05~\si{m}$ denoted by the red dotted line. After the collision, the norm of the relative velocity between the quadrotor and payload projected on the cable $\norm{\robotvel{k||}-\dot{\mathbf{p}}_{k||}}_2$ becomes $0$. Subsequently, the controller recovers the payload position and orientation to the desired pose. 
\subsubsection{Real-world vs. Simulation Comparison}
In this section, we compare the results from the real world and the simulation to prove the ability of our simulator to closely mimic real-world systems using our hybrid system collision model in the challenging case of an aerial transportation system made of three quadrotors. 

As shown in Fig.~\ref{fig:real-world-collide-demo-before-poking}, we hover three ``Dragonfly" quadrotors with a triangular suspended payload via 3 $1.05~\si{m}$ long cables. The controller used in real-world experiments is the same as the one we introduce in Section~\ref{sec:mult-cable-control}. We use a Vicon motion capture system to provide state feedback to the proposed controller and record the system states as well. While the system is hovering, we poke the payload's front corner with a long stick to make the payload pitch upward, leading the front cable to be slack as shown in Fig.~\ref{fig:real-world-collide-demo}. The cable becomes taut at a later stage after the quadrotors adjust themselves to make the cable taut and control the payload back to the desired hovering state. We then initialize the system in the simulator with the same parameters and the recorded states from Vicon in the corresponding real-world experiment when the front cable becomes slack. The results from the real-world experiment and simulation are shown in Figs.~\ref{fig:multi_robot_collide} and \ref{fig:multi-collide-payload-vel} - \ref{fig:multi-collide-payload-ang-vel}.

In Figs.~\ref{fig:multi-collide-payload-vel} and \ref{fig:multi-collide-payload-ang-vel}, we can observe how the payload's velocity and angular velocity changes when the front cable transitions from slack to taut in both simulation and real-world experiments. The red dash lines in both figures represent the moment when the front cable becomes taut from slack in the simulation. 

The plots show that the payload linear and angular velocity trends before and after the collision in both the real-world data and simulation data are closely aligned. Further error plots show that the errors between the real-world data and simulation data are bounded and converge to 0 in the end. We believe that an error within this range is acceptable for the purposes of our research and provides a substantial level of accuracy for users of our simulator. We also show in the attached multimedia material the entire testing procedure, additional experiments, and the collision simulation of multiple quadrotors carrying a rigid body payload where two cables transition from slack to taut.

\section{Discussion and Conclusion}\label{sec:conclusion}
In this paper, we proposed a simulator for aerial transportation and manipulation including single and multiple physically interconnected quadrotors via passive mechanisms such as cables and rigid links. This work marks the first instance of a full hybrid system dynamics model that incorporates slack and taut suspended cables between multiple quadrotors and payload. Our model includes a collision model for transient dynamics between two different system states, supported by a proven closed-form analytical solution.

Furthermore, we have developed trajectory planners and controllers for aerial manipulation with passive mechanisms and provided user-friendly interfaces to utilize these modules. The simulator also allows for the loading of a diverse range of system setups, making the simulator a versatile tool for both research and educational purposes. To foster innovation and collaboration within the community, we have made the simulator and its corresponding algorithms open-source.

We validated RotorTM's effectiveness and fidelity through approaches and the overall simulation framework with a series of real-world and simulated experiments, including trajectory tracking and hybrid dynamics. The results demonstrate the capability of our modeling, planning, and control approaches to accurately simulate hybrid dynamics and enable payloads to follow trajectories generated by our motion planner.

While our simulator has shown sufficient fidelity and provides flexible interfaces for researchers, we acknowledge that there are areas for further enhancement. Currently, the simulator does not include motor dynamics, and it does not account for external drag forces acting on the payload and the MAVs. Moreover, the current version is limited to quadrotors. In future iterations, we plan to incorporate low-level rotor dynamics and relevant rotor drag forces. We also aim to extend the range of aerial robots to include other types of aerial robots like hexacopters and octocopters. We want to stress that the proposed simulator should not be considered a monolithic solution, and we hope, by sharing this framework, to encourage the robotics community to collaborate and contribute by adding additional functionalities that may be helpful for different purposes.

Looking ahead, we envision endowing our simulator with photo-realistic features to support data collection for robotics and machine learning communities focused on learning and navigation for aerial transportation and manipulation. We also plan to extend the simulator to include active manipulation mechanisms, thereby broadening its appeal and usability within the community.

\bibliographystyle{IEEEtran}	
\bibliography{references}

\begin{IEEEbiography}[{\includegraphics[width=1in,height=1.5in,clip,keepaspectratio]{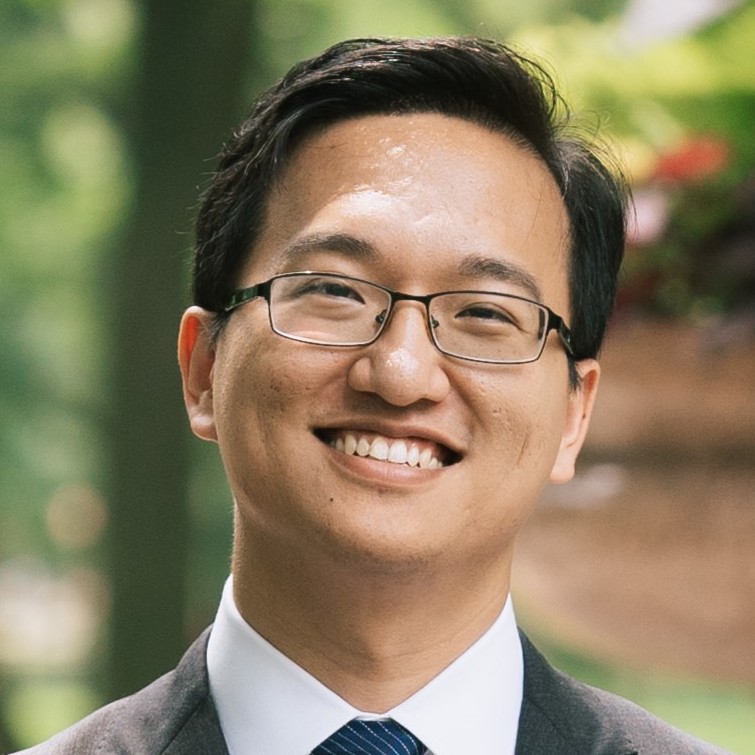}}]{Guanrui Li} 
earned his Ph.D. in Electrical and Computer Engineering at New York University, USA, with a focus on robotics and aerial systems. He obtained his Bachelor degree in Theoretical and Applied Mechanics from Sun Yat-sen University, where he was recognized as an Honors Undergraduate, and his Master degree in Robotics from the GRASP Lab at the University of Pennsylvania. His research is centered on the dynamics, planning, and control of robotics systems, with applications in aerial transportation and manipulation, as well as human-robot collaboration. Guanrui has received several notable recognitions, including the NSF CPS Rising Stars in 2023, the  Outstanding Deployed System Paper Award finalist at 2022 IEEE ICRA, and the 2022 Dante Youla Award for Graduate Research Excellence at NYU. He has an extensive publication record in top-tier robotics conferences and journals like ICRA, RA-L, and T-RO, and his work has garnered attention in various media, including IEEE Spectrum and the Discovery Channel.
\end{IEEEbiography}

\begin{IEEEbiography}[{\includegraphics[width=1in,height=1.5in,clip,keepaspectratio]{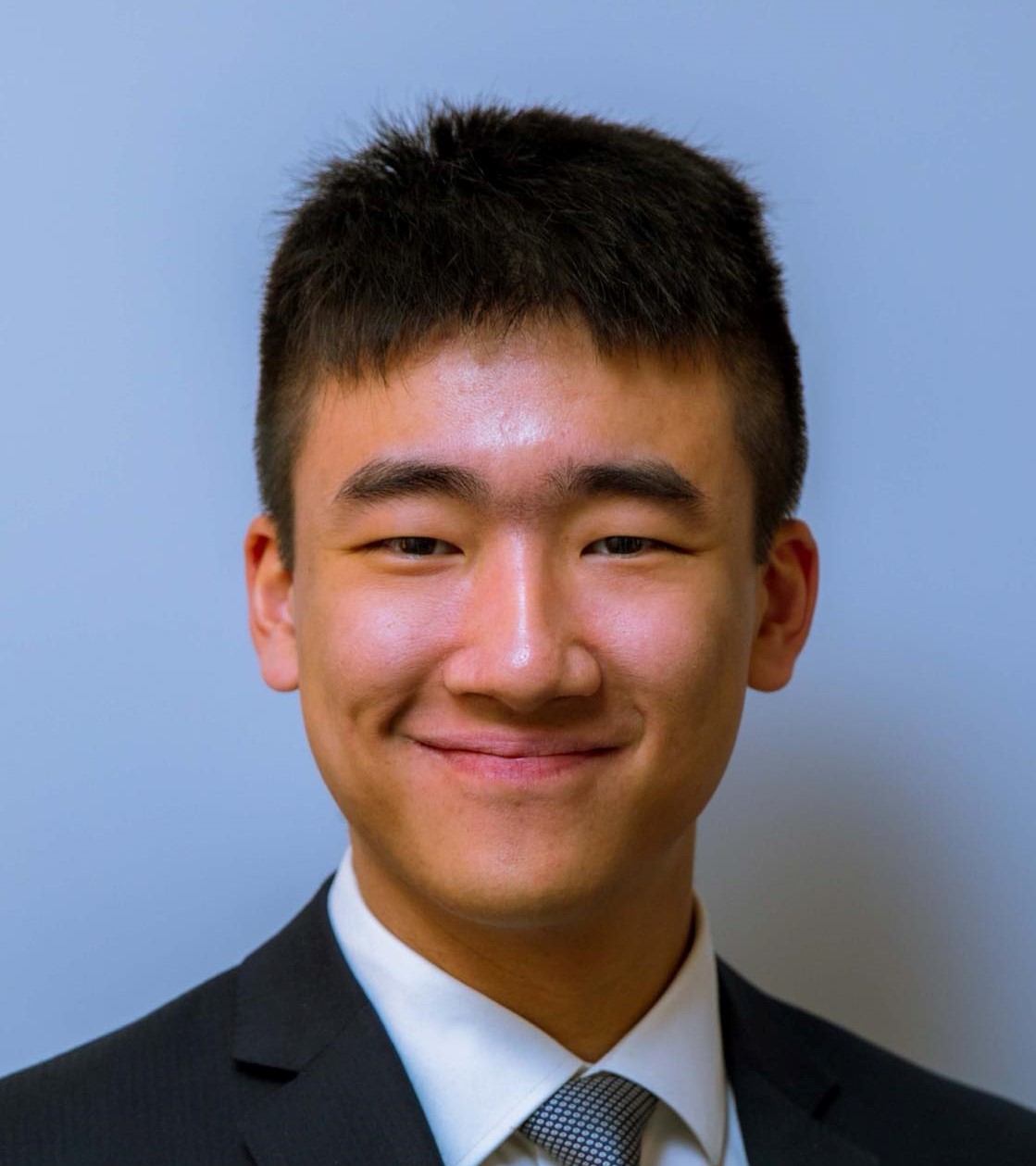}}]{Xinyang Liu}
Xinyang Liu was borned in Beijing, China, in 1999. He received his B.S. degree in mechanical engineering with double minors in robotics and computer science from New York University, New York, NY, in 2022, where he was named a University Honor Scholar and was a member of Tau Beta Pi. He is currently in his final year pursuing an M.S. degree in Aeronautics and Astronautics at Stanford University, Stanford, CA. His research interests include control theory, robotics, human-robot interactions, and autonomous systems.
\end{IEEEbiography}

\begin{IEEEbiography}[{\includegraphics[width=1in,height=1.5in,clip,keepaspectratio]{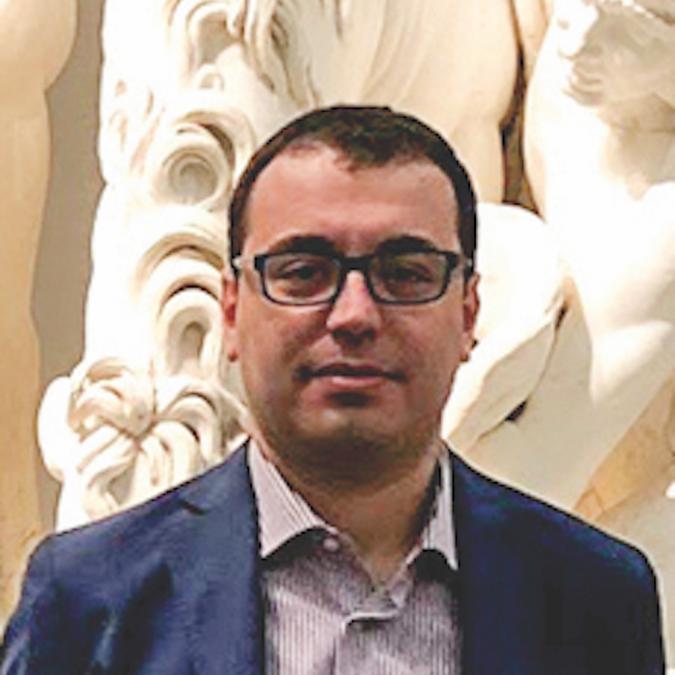}}]{Giuseppe Loianno} is an assistant professor at the New York University, USA and director of the Agile Robotics and Perception Lab (https://wp.nyu.e
du/arpl/) working on autonomous robots. He received a Ph.D. in robotics from University of Naples ”Federico II”, Italy in 2014. Prior joining NYU, he was post-doctoral researcher, research scientist and
team leader at the GRASP Lab at the University of
Pennsylvania in Philadelphia, USA. Dr. Loianno has
published more than 70 conference papers, journal
papers, and book chapters. His research interests
include perception, learning, and control for autonomous robots. He received
the NSF CAREER Award in 2022 and DARPA Young Faculty Award in
2022. He is recipient of the IROS Toshio Fukuda Young Professional Award
in 2022, Conference Editorial Board Best Associate Editor Award at ICRA
2022, Best Reviewer Award at ICRA 2016, and he was selected as Rising
Star in AI from KAUST in 2023. He is also currently the co-chair of the
IEEE RAS Technical Committee on Aerial Robotics and Unmanned Aerial
Vehicles. He was the the general chair of the IEEE International Symposium
on Safety, Security and Rescue Robotics (SSRR) in 2021 as well as program
chair in 2019, 2020, and 2022. His work has been featured in a large number
of renowned international news and magazines.

\end{IEEEbiography}
\vfill

\end{document}